\documentclass[10pt,twocolumn,letterpaper]{article}
\usepackage{minitoc}
\usepackage{iccv}
\usepackage{times}
\usepackage{epsfig}
\usepackage{graphicx}
\usepackage{amsmath}
\usepackage{amssymb}
\usepackage{booktabs}
\usepackage{cuted}
\usepackage{capt-of}
\usepackage{soul}
\makeatletter
\@namedef{ver@everyshi.sty}{}
\makeatother
\usepackage{tikz}
\usepackage[toc,page,header]{appendix}
\usepackage{minitoc}
\usepackage{pifont}
\usepackage[pagebackref=true,breaklinks=true,colorlinks,bookmarks=false]{hyperref}
\usepackage[capitalize]{cleveref}
\usepackage{xcolor,colortbl}
\usepackage{multirow}
\usepackage{tikz}
\usepackage{bbding} 
\usepackage{lipsum}
\usepackage{stfloats}
\usepackage[weather]{ifsym}
\newcommand{\ind}{\perp\!\!\!\!\perp}
\newcommand{\red}[1]{\textcolor{red}{#1}}
\definecolor{emerald}{RGB}{80,200,120}
\usepackage{float}

\newcommand{\green}[1]{\textcolor{emerald}{#1}}

\newcommand{\gray}[1]{\textcolor{gray}{#1}}
\definecolor{lightgray}{rgb}{0.83, 0.83, 0.83}
\definecolor{Gray}{gray}{0.6}
\definecolor{aliceblue}{rgb}{0.94, 0.97, 1.0}
\definecolor{mistyrose}{rgb}{1.0, 0.89, 0.88}

\usepackage[capitalize]{cleveref}
\crefname{section}{Sec.}{Secs.}
\Crefname{section}{Section}{Sections}
\Crefname{table}{Table}{Tables}
\crefname{table}{Tab.}{Tabs.}

\def\OURS{VFC\xspace}

\iccvfinalcopy 

\def\iccvPaperID{4413}

\ificcvfinal\fi

\begin{document}
\doparttoc 
\faketableofcontents

\title{Verbs in Action: Improving verb understanding in video-language models}

\author{Liliane Momeni$^{1}$ \quad Mathilde Caron$^{2}$ \quad Arsha Nagrani$^{2}$ \quad Andrew Zisserman$^{1}$  \quad Cordelia Schmid$^{2}$ \\
$^{1}$ Visual Geometry Group, University of Oxford, UK \\
$^{2}$ Google Research \\
}

\maketitle
\ificcvfinal\thispagestyle{empty}\fi

\begin{strip}\centering
\vspace{-10mm}
\includegraphics[width=\textwidth]{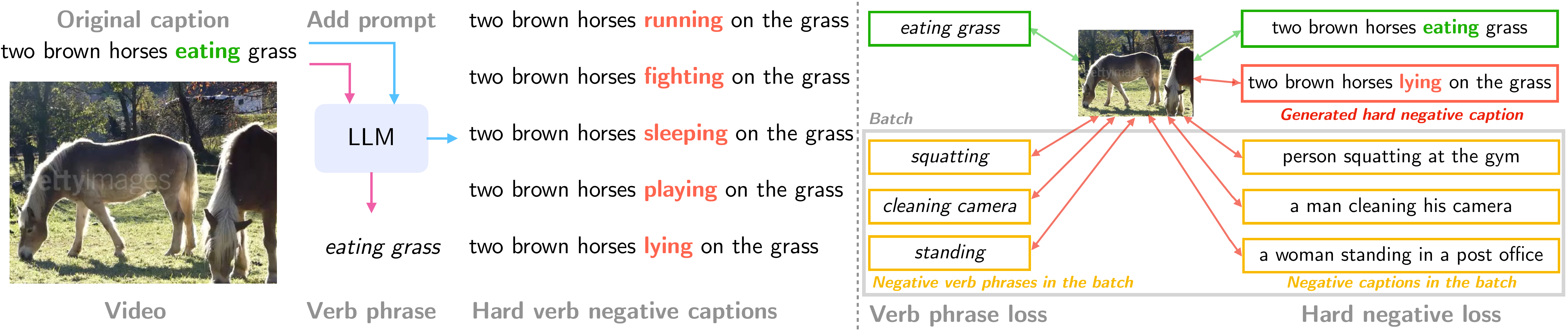}
\mbox{}\\
\captionof{figure}{\textbf{Verb-Focused Contrastive (\OURS) learning}: (Left): Given a video and its corresponding caption, we leverage a Large Language Model (LLM) to output (1)~hard negative captions, where only the verb has been changed while keeping the remaining context, and (2)~verb phrases which succinctly describe the action in the video. (Right): To encourage better verb reasoning, we subsequently enforce (1)~a \textit{calibrated} hard negative loss, using our generated hard negative captions and other captions in the batch, and (2)~a fine-grained, verb phrase loss.
We show that \OURS improves verb understanding of video-language models compared to the standard contrastive loss. 
\label{fig:teaser}}
\end{strip}

\begin{abstract}
Understanding verbs is crucial to modelling how people and objects interact with each other and the environment through space and time. Recently, state-of-the-art video-language models based on CLIP have been shown to have limited verb understanding and to rely extensively on nouns, restricting their performance in real-world video applications that require action and temporal understanding. In this work, we improve verb understanding for CLIP-based video-language models by proposing a new Verb-Focused Contrastive (VFC) framework. This consists of two main components: (1) leveraging pretrained large language models (LLMs) to create hard negatives for cross-modal contrastive learning, together with a calibration strategy to balance the occurrence of concepts in positive and negative pairs; and (2) enforcing a fine-grained, verb phrase alignment loss. Our method achieves state-of-the-art results for \textit{zero-shot} performance on three downstream tasks that focus on verb understanding: video-text matching, video question-answering and video classification. To the best of our knowledge, this is the first work which proposes a method to alleviate the verb understanding problem, and does not simply highlight it.
\end{abstract}
\section{Introduction}\label{sec:intro}

Large-scale visual-language models (VLMs) such as CLIP~\cite{Radford2021CLIP} have shown strong performance on multiple video-language tasks such as text-to-video retrieval~\cite{Luo2021CLIP4Clip}, video question-answering, and open-set action recognition~\cite{lin2022frozen}. These models perform surprisingly well on these tasks in a zero-shot setting, despite being trained only on image-language pairs (with no access to temporal data), even outperforming strong video-specific models~\cite{bain2021frozen,yan2022multiview}.

A recently highlighted and well-documented problem with such models, however, is their strong \textit{noun} or \textit{object} bias, as evidenced by their lower performance in distinguishing between \textit{verbs} in natural language descriptions~\cite{hendricks2021probing,park-etal-2022-exposing,2210.01936}. This was first studied in images alone by the SVO-Probes benchmark~\cite{hendricks2021probing}, which shows that \textit{image}-language models struggle to distinguish between different verbs, and often rely on the nouns instead. This problem persists with \textit{video}-language models that inherit these VLMs, even after they are fine-tuned on video-text datasets~\cite{7780940,lsmdc}. For example, Park {\it et al}.~\cite{park-etal-2022-exposing} similarly propose evaluation sets with hard verb negatives, and show that CLIP-based models, even when fine-tuned on video datasets, have difficulties discriminating verbs in a multi-choice setting where the context remains unchanged. Yuksekgonul {\it et al}.~\cite{2210.01936} further highlight limitations of vision-language models at understanding attribute, relationship, and order information. This deficiency in verb understanding limits the model's applicability for real-world tasks. Verbs encapsulate how people and objects interact with each other, and the environment, via actions in space and time. 

 We believe that there are two probable causes for this deficiency, even after fine-tuning on video-text data: (i) existing visual-text datasets have a strong bias towards single-frame concepts such as \textit{objects} and \textit{backgrounds} as well as \textit{static} actions~\cite{SevillaLara2021OnlyTC, buch2022revisiting,revealing_single}. Models are hence less incentivized to understand dynamics and temporal actions~\cite{SevillaLara2021OnlyTC}, biasing them towards noun understanding; and (ii) the limitations of the cross-modal contrastive pretraining objective used by most current vision-language models~\cite{2210.01936}. In contrastive learning, the model is trained to distinguish correct video-caption pairs from incorrect ones. Since it is unlikely that existing datasets contain many examples with captions of \textit{similar} context but \textit{different} verbs, the task can be solved by taking little verb information into account. This relates to shortcut learning in deep neural networks~\cite{Geirhos_2020}.

In an attempt to mitigate this problem, we propose a novel training framework for tackling the task of verb understanding in vision-language models. Our framework, called  \textbf{V}erb-\textbf{F}ocused \textbf{C}ontrastive pretraining (VFC), consists of two novel technical modifications to the contrastive learning framework. We first introduce a method to automatically generate negative sentences for training where only the verb has changed, keeping the context the same. This is done using LLMs~\cite{2020t5, palm}, in an automatic and scalable manner. Note that we \textit{generate} hard negative captions, unlike works that simply mine hard negatives from an existing paired dataset~\cite{rdk+23}, or change the order of words~\cite{2210.01936}.  For example, given the caption `\textit{two brown horses eating grass}', we generate the negative caption `\textit{two brown horses running on the grass}' (see Fig.~\ref{fig:teaser}).
While this improves performance on some downstream tasks, we find that introducing concepts simply in \textit{negative} examples can also lead to an imbalance in the contrastive objective, favouring certain concepts in the feature space. To solve this, we propose a simple but effective \textit{calibration strategy} to balance the occurrence of verbs in both positive and negative captions.

Secondly, inspired by recent works on \textit{grounding} concepts in vision-language learning~\cite{kamath2021mdetr,cao2022locvtp}, we also introduce a verb phrase loss that explicitly isolates the verb from a caption for more focused training. For example, we extract the verb phrase `\textit{eating grass}' from the caption `\textit{two brown horses eating grass}' (see Fig.~\ref{fig:teaser}).  We find that this helps particularly for zero-shot performance on downstream tasks that do not use long sentences in their evaluation~\cite{Ghadiyaram2019LargeScaleWP}. Verb phrases are also extracted from sentences using LLMs.

We then train a CLIP-based model~\cite{Luo2021CLIP4Clip} on a video-language dataset with this novel training framework. We show that a \textit{single model} trained in this way transfers well to diverse downstream tasks that focus particularly on verb understanding, including three video benchmarks  (multiple choice video-text matching on MSR-VTT~\cite{7780940}, video question answering on Next-QA~\cite{xiao2021next}, action recognition on Kinetics~\cite{Carreira_2017_CVPR}) and one image benchmark (SVO-probes~\cite{hendricks2021probing}), achieving state-of-the-art performance compared to previous works in \textit{zero-shot} settings (and often with fine-tuning as well); while maintaining performance on noun-focused settings. On Kinetics, we also introduce a verb split of the data which specifically highlights classes that are challenging to distinguish without fine-grained verb understanding (`\textit{brushing hair}' vs `\textit{curling hair}') and show that our model particularly improves performance on this split.

\section{Related works}\label{sec:relatedwork}
\vspace{-0.1cm}

\noindent\textbf{LLMs for video-text tasks.} 
LLMs have been used for various vision applications, for example to initialise vision-text models~\cite{mvgp, Chen_2022_CVPR, vc_gpt}. Recent works further use frozen LLMs via prompting for tackling vision-language tasks~\cite{flamingo, tsimpoukelli2021multimodal, zeng2022socraticmodels, yang2022frozenbilm, wang2022language, yang2021empirical, chen2023see}. LLMs have also been used in creative ways to obtain better supervision for training for various tasks~\cite{yang2021justask, zellers2022merlotreserve, lin2022learning, zhao2022lavila, athousandwords}. For example, \cite{yang2021justask} use LLMs to generate question-answer pairs from transcribed video narrations, while \cite{zellers2022merlotreserve} use LLMs to rephrase questions into sentences. \cite{lin2022learning} use LLMs to match noisy speech transcriptions to step descriptions of procedural activities. \cite{nagrani2020speech2action} train BERT~\cite{devlin-etal-2019-bert} to predict action labels from transcribed speech segments and use this to scale up training data for action classification. \cite{zhao2022lavila} use pretrained LLMs conditioned on video to create automatic narrations. Recent works~\cite{zhao2022lavila, athousandwords} also show the benefits of using LLMs to paraphrase captions for data augmentation for video-language pretraining. \cite{Li2022CLIPEventCT} use LLMs to generate negative captions by manipulating event structures. Our work differs to~\cite{Li2022CLIPEventCT} in that we focus specifically on verb negatives, and videos instead of images.  Most closely related to our work, \cite{park-etal-2022-exposing} construct a test set for verb understanding by leveraging T5~\cite{2020t5} and highlight the poor performance of current video-language models. Our work is substantially different: (i) we automatically construct hard negative captions for \textit{training} (not testing), (ii) we compare the use of different LLMs, (iii) we show that training with such negative captions can improve verb understanding on various verb-focused benchmarks.

\noindent\textbf{Hard negatives for contrastive pretraining.} 
Hard negatives have been used to improve performance in metric representation learning and contrastive learning~\cite{Kalantidis,Harwood,Wu}. Recent works mine hard negatives from an existing paired dataset~\cite{rdk+23,xu-etal-2021-videoclip,Yang2021TACoTC}. In comparison, in our work, we \textit{generate} hard negative captions and propose a careful calibration mechanism for training effectively with such unpaired data. We also verify here the benefit of the HardNeg-NCE loss~\cite{rdk+23} when training with generated hard negative captions. \cite{2210.01936} construct hard negative captions by shuffling words from the original caption to improve order and compositionality understanding. Our work differs by (i)~focusing specifically on \textit{verb} reasoning, as opposed to object-attribute relationships, (ii)~using LLMs to construct hard verb text negatives as opposed to perturbing the word order, (iii)~focusing on \textit{video}-language models.

\noindent\textbf{Learning from parts-of-speech in video.} Recent works use parts-of-speech (PoS) tags for video understanding~\cite{Sadhu_2021_CVPR, wray2019learning, augmentation, Ghadiyaram2019LargeScaleWP,Xu2015JointlyMD}. \cite{wray2019learning} learn multi-label verb-only representations, while other works focus on learning adverb representations~\cite{Doughty2019ActionML,doughty2022how}. \cite{Alayrac16unsupervised} use verb-noun pairs for unsupervised learning with instructional videos, while \cite{augmentation} leverage such pairs to generate data augmentations in the feature space. Other works exploit PoS for fine-grained or hierarchical alignment between video and text~\cite{bowenzhang, chen2020fine}. \cite{wray2} learn a separate multi-modal embedding space for each PoS tag and then combine these embeddings for fine-grained action retrieval. \cite{chen2020fine} construct a hierarchical semantic graph and use graph reasoning for local-global alignments. Most closely related to our work, \cite{Yang2021TACoTC} use a PoS based token contrastive loss. Our work differs in that: (i) we apply a verb phrase contrastive loss, as opposed to separate verb and noun losses; (ii) we extract verb phrases using a LLM and show this performs better than PoS tagging with NLTK~\cite{bird2009natural} (Tab.~\ref{fig:multi_ablat}); (iii) we evaluate our methods on verb-focused downstream tasks. Similarly to~\cite{Ghadiyaram2019LargeScaleWP}, we find that training with verb phrase supervision helps for zero-shot performance on tasks with shorter sentences.

\noindent\textbf{Temporal understanding in videos.} 
A long term goal in computer vision is temporal understanding in videos~\cite{Carreira_2017_CVPR,ssv2, Diba2019LargeSH,howmany, lfb2019,zhou2017temporalrelation, whatactions}. However, current training and test datasets have a strong visual bias towards \textit{objects} and \textit{backgrounds} as well as \textit{static} actions~\cite{SevillaLara2021OnlyTC,whatmakesavideo}, with some works~\cite{buch2022revisiting, revealing_single} demonstrating strong results with a \textit{single} frame. Despite these challenges, many recent works in video-only self-supervised learning propose pretext tasks for improving temporal modelling~\cite{Kim_2019,VideoJigsaw,arrow1, wei2018learning, Price2019RetroActionsL,speednet,Wang20,Yao2020VideoPR, Dorkenwald2022,Misra2016ShuffleAL, Liang2021SelfsupervisedSR, wang, Behrmann2021LongSV,broaden,dave2022tclr}. Unlike these works that use only video, ~\cite{longform,cao2022locvtp} focus on fine-grained temporal video-text alignment via localization of text sub-tokens. \cite{instillingtime} also leverage before/after relations in captions to create artifical training samples for video-text. Differently to these works (which create augmented video negatives or positives), we approach the problem of improving \textit{verb understanding} in video-language models from the language side, by leverging the strong generalization capabilities of LLMs.

\section{Method}\label{sec:method}

\begin{figure*}[t]
    \centering
    \includegraphics[width=\textwidth]{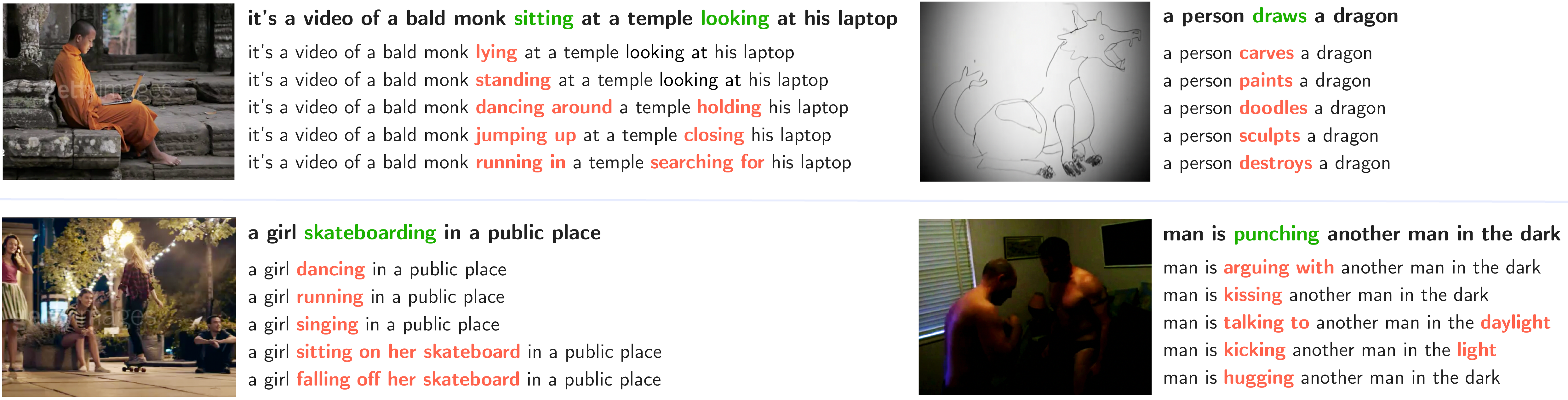}
        \vspace{-0.2cm}
    \caption{\textbf{Qualitative examples of hard negatives generated by PaLM.} We show a single frame per video and the corresponding caption in bold, with the verb highlighted in green. We see that PaLM can effectively generate hard negatives where the verb has changed (changes in red).
    When there are several verbs in the caption (see top left), PaLM may replace one or all verbs.
    As a failure case (bottom right), we show an example where PaLM can change more than just the verb, which could make it an easier negative (replacing `punching' by `talking' but also `dark' by `daylight'). 
    }
    \label{fig:qualitative-main}
    \vspace{-0.4cm}
\end{figure*}

Our goal is to adapt large-scale vision-language pretrained models (such as CLIP) to understand \textit{verbs}. We aim to do this without requiring such models to be retrained from scratch, but by simply fine-tuning them on a video-language dataset.
However, given the pitfalls with using the standard video-text contrastive setup~\cite{Radford2021CLIP} on existing video-language datasets, we propose a new framework which we call \textbf{V}erb-\textbf{F}ocused \textbf{C}ontrastive pretraining (\OURS).
It consists of two components, both using the power of LLMs:
(i)~a novel calibrated hard negative training method where we train with synthetic verb-focused hard negative captions, and
(ii)~an additional verb phrase loss where videos are contrasted against isolated verb phrases as opposed to the entire caption.
Note that a `verb phrase' can be a single verb or verb-noun pair depending on the caption (see Fig.~\ref{fig:teaser}).

\subsection{Preliminaries}\label{sec:prelim}

\noindent\textbf{Large Language Models (LLMs)} are generative text models with impressive capacities, in particular for few-shot or prompt-based learning~\cite{palm}. In our work, we design prompts to instruct a LLM to (i)~create verb-focused hard negative captions and (ii)~isolate verb phrases from the captions of a dataset.
LLMs allow scalability and generalisation, and as we show in the ablations (see Tab.~\ref{tab:hn-generation}~and Tab.~\ref{fig:multi_ablat}), are preferable to manual or rule based methods (eg.~NLTK~\cite{bird2009natural}). In particular, we use PaLM~\cite{palm}, a state-of-art autoregressive model, throughout this paper.
However, our framework is agnostic to this choice and other LLMs can be used instead (see Tab.~\ref{tab:hn-generation}).

\noindent\textbf{Video-language contrastive pretraining} works by learning to distinguish between aligned and non-aligned video-text pairs.
Given a dataset of $N$ pairs $\{(V_i, T_i)\}_{i \in N}$ with video $V_i$ and caption text $T_i$, we extract normalised feature representations $v_i$ and $t_i$ by using a video encoder $f$ and text encoder $g$: we have $v_i = f(V_i)$ and $t_i=g(T_i)$.
We use the InfoNCE loss~\cite{infonce} to make aligned (`positive') pairs close in feature space and all other pairwise combinations in the batch further apart~\cite{Radford2021CLIP}.
We optimize for video-to-text $L^{v2t}$ and text-to-video $L^{t2v}$  alignments:
\vspace{-0.3cm}
\begin{equation}
\vspace{-0.3cm}
\label{eq:loss1}
    L^{t2v}_i = - t_i^\top v_i / \sigma + \log\sum_{j=1}^{B} \exp(t_i^\top v_j / \sigma)
\end{equation}
where $B$ is the batch size and $\sigma$ a temperature parameter controlling the sharpness of the distribution.
$L^{v2t}$ is obtained by inverting $v$ and $t$ in Eq.~\ref{eq:loss1}.

\noindent\textbf{Adapting image-text models to videos.}
We leverage CLIP~\cite{Radford2021CLIP} for video-language tasks following the CLIP4CLIP `seqTrans' protocol~\cite{Luo2021CLIP4Clip}. Both single-modal encoders (video $f$ and text $g$) are initialized with CLIP weights, with four additional temporal frame aggregation transformer blocks stacked on top of the image encoder (see~\cite{Luo2021CLIP4Clip} for more details). Our approach is agnostic to model architecture and so any state-of-the-art video-language architecture could be potentially used.

\subsection{Verb-Focused Contrastive Pretraining (VFC)} \label{sec:vfc}

We describe both our calibrated hard negative training (Sec.~\ref{subsection:neg-methods}) and the proposed verb phrase loss (Sec.~\ref{subsection:met:verbloss}).

\subsubsection{Calibrated Hard Negative training}\label{subsection:neg-methods}

In regular contrastive learning, given a video-caption pair, other captions in the batch are simply pushed further in the feature space.
Since it is unlikely that existing datasets contain many examples with captions of similar context but different \textit{verbs}, the task can be solved by paying little attention to verbs.
Instead, our goal is to encourage the video-language model to focus on verb reasoning.
We do so by tasking a LLM to generate hard negative captions where only the verb(s) in the captions change.
Second, we train with these additional negative captions. 
We find that naive training with additional data leads to imbalances affecting the resulting video-text feature space. We propose a simple but effective calibration mechanism to solve this.

\noindent\textbf{Generating verb-focused hard negatives with PaLM.}
Given a caption $T_i$, we task PaLM to replace the verbs with other verbs that convey a different action, but still form a linguistically and semantically viable sentence (which may not be guaranteed with random verb replacements -- see qualitative examples in Sec.~\ref{sec:app:palm-comparison-neg} of the appendix). 
For example, in the caption `\textit{a man washes his face}', the verb `\textit{washes}' should not be replaced with `\textit{jumps}' or `\textit{plays}'. The generated caption is then a negative match for the corresponding video $V_i$ (albeit a \textit{hard} negative, as the nouns and context remain the same).
We experiment with different handcrafted prompts, and find our best performing prompt to be the following: \textit{`In this task, you are given an input sentence. Your job is to tell me 10 output sentences with a different meaning by only changing the action verbs'}. We also add four input-output pair examples to the prompt, which increases the quality of PaLM's predictions (see Sec.~\ref{subsec:input-output} of the appendix).
We use one PaLM forward pass per caption $T_i$ to generate ten verb-focused hard negatives for that caption (qualitative examples of the generated captions can be seen in Fig.~\ref{fig:qualitative-main}).
During training, we randomly sample $N^{\text{hard}}$ generated captions for each pair $(V_i, T_i)$ in the minibatch, which we denote $\left(T^{\text{hard}}_{i_k}\right)_{k \in [1, N^{\text{hard}}]}$.
Importantly, note that a $T^{\text{hard}}_{i_k}$ is a new generated text caption, or an \textit{unpaired} data sample, meaning that it does not come with a corresponding matching (`positive') video.

\begin{table*}[t]
\small
\setlength{\tabcolsep}{2pt}
\centering
\begin{tabular}{l|c|l}
\toprule
Name &  Video-to-text alignment loss & \multicolumn{1}{c}{$R_\omega$} \\
\midrule
Baseline  & $ - v_i^\top t_i / \sigma + \log\sum_{j=1}^{B} \exp(v_i^\top t_j / \sigma)$ & $\quad \frac{(B - 1)S_\omega}{S_\omega} \qquad \quad~~ \ind \omega $ \\

HN  & $ - v_i^\top t_i / \sigma + \log\left(\sum_{j=1}^{B} \exp(v_i^\top t_j / \sigma) + \sum_{j=1}^{B} \sum_{k=1}^{N^{\text{hard}}}\exp(v_i^\top t^{\text{hard}}_{j_k} / \sigma) \right)$  & $\quad \frac{(B - 1)S_\omega + BG_\omega}{S_\omega} ~~~~ \propto B\frac{G_\omega}{S_\omega}$  \\

 \rowcolor{aliceblue} Calibrated HN  & $- v_i^\top t_i / \sigma + \log\left(\sum_{j=1}^{B} \exp(v_i^\top t_j / \sigma) + \sum_{k=1}^{N^{\text{hard}}}\exp(v_i^\top t^{\text{hard}}_{i_k} / \sigma) \right)$ & $\quad \frac{(B - 1)S_\omega + G_\omega}{S_\omega} ~~~~~~ \propto \frac{G_\omega}{S_\omega}$ \footnotesize{with} $G_\omega \thickapprox S_\omega$ \\
\bottomrule
\end{tabular}
\vspace{0.2cm}
\caption{\textbf{Different choices for video-to-text alignment} when training with additional hard negatives (HN).
$R_\omega$ is the ratio of the number of times a given verb phrase $\omega$ is used as a negative versus the number of times it is used as a positive. We note that for the regular contrastive loss (\textbf{Baseline}), $R_\omega$ only depends on the batch size $B$, however when training with generated hard negatives (\textbf{HN}), it depends on the verb phrase $\omega$. We minimise this effect using our proposed \textbf{Calibrated HN} loss, which we denote as $L^{\text{CHN}}_{i}$. See details in Section~\ref{subsection:neg-methods}.
}
\label{tab:ratios}
\vspace{-0.4cm}
\end{table*}

\noindent\textbf{Calibration.}
Interestingly, we observe that naively adding in negative captions into training with a contrastive loss leads to harmful feature space distortions, as some concepts are only seen in negative captions but never in positives. 
This is observed by careful analysis of downstream performance (see study in Tab.~\ref{tab:naive-hn}~and Tab.~\ref{tab:calibration}). We hence next describe a calibration mechanism to avoid such distortions:
we first denote the vocabulary of all verb phrases in the original and generated captions as $\Omega$.
For each verb phrase $\omega$ (or `concept') in $\Omega$, we use $S_{\omega}$ to represent the number of times it appears in the captions of the original dataset and $G_{\omega}$ for the number of times it appears in the PaLM-generated captions.
We then derive equations for $R_\omega$ (see Tab.~\ref{tab:ratios}), which we define as the ratio of the number of times a verb phrase $\omega$ is used as a negative versus as a positive during training, for different choices of the video-to-text contrastive loss (note $L^{t2v}$ is unchanged).

\noindent \textbf{Contrastive training with paired data (Baseline).}
We first note that \textit{the ratio $R_\omega$ is independent of the verb phrase $\omega$} in regular contrastive learning (paired data only).
It simply depends on the batch size $B$, as $S_w$ is cancelled from both the numerator and denominator.
This means that the number of times a concept is used as a positive versus negative sample is the same regardless of the considered verb phrase.
This naturally balances training, and is a great property of the contrastive framework. \\
\textbf{Adding generated unpaired negative captions (HN).} 
However, when training with unpaired captions, this ratio is proportional to $G_\omega / S_\omega$ and therefore becomes \textit{dependent} on the considered verb phrase $\omega$.
This can have significant consequences for the video-text feature representations. The model can learn to either ignore or always predict some concepts based on the average concept occurrences in positive or negative pairs during training.

\noindent \textbf{Hard negatives with calibration (Calibrated HN).} In order to make $R_\omega$ as $\omega$-agnostic as possible, we introduce an ensemble of two techniques which we refer to as `calibration'.
First, we ignore the hard negative captions from the other elements of the batch (see row 3 in Tab.~\ref{tab:ratios}), which allows us to mitigate the influence of $G_\omega / S_\omega$ by not amplifying it by the batch size $B$ (equal to 256).
Second, we filter the generated PaLM captions to have $G_{\omega} \thickapprox  S_{\omega}$.
In practice, we discard some generations so that the number of times a verb phrase appears in the set of kept generations is equal to the number of times it is originally present in the dataset.
We denote our video-to-text loss (text-to-video is unchanged) as $L^{\text{CHN}}_{i}$ for calibrated hard negative training.

\noindent\textbf{Video mining.}
An alternative to avoid imbalances due to the addition of negative captions would be to avoid training with unpaired data at all, by mining a matching video $V^{\text{hard}}_{i_k}$ for each generated caption $T^{\text{hard}}_{i_k}$.
We attempt this via CLIP-based text-to-video retrieval in a large video database but found that
finding a video matching a detailed, long caption is challenging, as such a precise video may not exist in a given corpus (see Sec.~\ref{subsec:video-mining} in the appendix for examples).

\vspace{-0.3cm}
\subsubsection{The verb phrase loss}\label{subsection:met:verbloss}
\vspace{-0.1cm}
In order to further encourage our model to focus on verbs, we introduce a contrastive `verb phrase' loss.
We use PaLM to extract the verb phrase $T^\text{verb}_i$ in a caption $T_i$ with the following prompt: `\textit{In this task, you are given an input sentence. Your job is to output the action verb phrases.}' While multiple parts-of-speech (PoS) tagging tools exist, we use a LLM for the following reasons:  (i) we would like to isolate verb phrases, which may correspond to single verbs or verb-noun pairs depending on the caption , (ii) LLMs deal better with ambiguous cases (see qualitative examples in Sec.~\ref{sec:app:palm-comparison-vp} of the appendix).
We show the benefits experimentally via an ablation in Tab.~\ref{fig:multi_ablat}. During training, we minimize the following loss:
\vspace{-0.3cm}
\begin{equation*}
\vspace{-0.3cm}
L^{\text{verb-phrase}}_i = - v_i^\top t^{\text{verb}}_i / \sigma + \log\sum_{j=1}^{B} \exp(v_i^\top t^{\text{verb}}_j / \sigma)
\end{equation*}
where the negative verb phrase representations $t^{\text{verb}}_j$ simply come from other captions in the batch.
Note that we do not require the calibration mechanism described in Section~\ref{subsection:neg-methods} since all verb phrases $T^\text{verb}_i$ have a positive video match $V_i$ (i.e. the video aligned with $T_i$). \\
Overall, our verb-focused contrastive (\OURS) pretraining optimizes the sum of three objectives:
\vspace{-0.3cm}
\begin{equation*}
\vspace{-0.3cm}
L^{\text{VFC}} = \frac{1}{B} \sum_{i=1}^{B} \left( \lambda_1 L^{t2v}_i + \lambda_2 L^{\text{CHN}}_{i} + \lambda_3 L^{\text{verb-phrase}}_i \right)
\end{equation*}
with parameters $\lambda_1$, $\lambda_2$ and $\lambda_3$ weighting the contribution of the different terms. We learn the parameters of $f$ and $g$ via back-propagation.

\subsection{Implementation details}\label{subsec:implementation}
\vspace{-0.1cm}

\noindent\textbf{Spoken Moments in Time (SMiT) pretraining dataset.}
The SMiT~\cite{Monfort_2021_CVPR} training set consists of 481K pairs of 3 seconds video clips with corresponding captions.
It is a subset of Moments in Time (MiT)~\cite{monfortmoments}.
Our work falls under the umbrella of transfer learning:
we pretrain on SMiT and then use the resulting features to solve different downstream tasks in a zero-shot or fine-tuned manner.
Pretraining is either done as in regular contrastive learning (`baseline') or with our \OURS framework.
We find that the baseline already performs competitively on our benchmarks, despite the relatively small size of SMiT compared to other datasets such as HowTo100M~\cite{miech2019howto100m}, due to the quality and diversity of the manually annotated captions.
We encourage the community to consider SMiT as a powerful pretraining dataset.

\noindent\textbf{PaLM.}
We use PaLM-540B~\cite{palm} with beam size 4, output sequence length 512, and temperature of 0.7. The negative captions are generated in an autogressive way and are therefore of arbitrary length.
We post-process them by removing text after any newline character and by filtering out candidates which contain the same verbs as the original caption.

\noindent\textbf{Training details.}~Most hyper-parameters follow CLIP4CLIP~\cite{Luo2021CLIP4Clip}.
We initialise our model with CLIP ViT/B-32 and train with \OURS for 100 epochs with a batch size of 256, base learning rate of 1e-7, weight decay of 1e-2, temperature of 5e-3 and weights $\lambda_1=2$, $\lambda_2=\lambda_3=1$ which we empirically find to work well in our experiments. Indeed, this balances the video-to-text and text-to-video loss terms. We also normalise each loss term by its value obtained from a random uniform prediction in order to have all loss terms in the same range (loss always equal to 1 for a random uniform prediction).
We sample 32 frames per video at 25fps, with a 2 frame stride.
See Sec.~\ref{sec:app:details} in the appendix for further implementation details and extensive evaluation protocols.

\section{Experiments}\label{sec:experiments}

We curate a suite of benchmarks from existing works to evaluate verb understanding which we present in Sec.~\ref{subsec:benchmarks}.
Then we ablate various components of our \OURS framework in Sec.~\ref{subsec:ablation}. 
Finally, we demonstrate improved performance on our diverse set of downstream tasks in Sec.~\ref{subsec:results}, and compare to the state of the art.

\begin{table}[t]
\small
\setlength{\tabcolsep}{6pt}
\centering
\begin{tabular}{lcll}
\toprule
Method & Hard negatives & ~~~~Verb$_{H}$ & ~~~~K-400  \\
\midrule
Baseline & $\varnothing$ & 69.9 & 55.6 \\
\midrule
\textit{w/o LLM} \\
         &  Random verb & 73.6 \green{\small{(+3.7)}} & 55.0 \red{\small{(-0.6)}}\\
         &   Antonym verb &  72.4 \green{\small{(+2.5)}} & 55.4 \red{\small{(-0.2)}} \\
\midrule
\textit{w/ LLM} \\
         &   T5~\cite{2020t5} & 75.1 \green{\small{(+5.2)}} &  55.8 \green{\small{(+0.2)}} \\
  \rowcolor{aliceblue}      Ours & PaLM~\cite{palm} &   78.0 \green{\small{(+8.1)}}  & 55.8 \green{\small{(+0.2)}}\\
\bottomrule
\end{tabular}
 \vspace{0.2cm}
    \caption{
    \textbf{Hard negatives generation.}
We explore both LLM based and non LLM-based methods to obtain hard negative captions.
Although PaLM LLM captions achieve the best performance, other LLMs (T5) achieve good results too.
All methods are evaluated with calibration.
    }
    \label{tab:hn-generation}
    \vspace{-0.2cm}
\end{table}

\subsection{Verb-Focused Benchmarks}\label{subsec:benchmarks}

\noindent\textbf{MSR-VTT multiple choice (MC)} is a benchmark of 10K videos of length 10--30 secs.
We evaluate on the standard 3k split and on Verb$_{H}$ from~\cite{park-etal-2022-exposing}.
In this setting, the task is to associate each video to the right caption among five choices.
While the four wrong captions are randomly chosen from other videos in the standard 3k split, one of them is replaced by a \textit{hard verb negative} in Verb$_{H}$~\cite{park-etal-2022-exposing}.

\noindent\textbf{Video question answering on NEXT-QA}
The train (resp. val) split contains 3870 (resp. 570) videos with 32K (resp. 5k) questions.
There are three types of questions: causal (C), temporal~(T) and descriptive~(D).
We consider the standard setting as well as ATP$_{hard}$~\cite{buch2022revisiting}, a subset automatically constructed with questions that are non-trivially solved with a single frame.
ATP$_{hard}$ is designed to be a better benchmark for the model's true causal and temporal understanding which we believe is strongly related to verb reasoning.

\noindent\textbf{Kinetics-400} is a video classification dataset with 400 human action classes.
We report top-1, top-5 and their average classification accuracy.
We follow~\cite{Radford2021CLIP} to evaluate classification in an open-set, zero-shot manner.
This benchmark allows to assess transfer ability to \textit{action} classification, which requires strong verb understanding  (given actions are usually described with verb phrases).

\noindent\textbf{SVO-probes dataset} is a benchmark specifically designed to measure progress in verb understanding of image-text models~\cite{hendricks2021probing}.
It contains image–caption pairs with 421 different verbs. We simply replicate the image multiple times as input to our video model.
We report Average Precision (AP) on the entire dataset as well as the verb-focused setting (details about our evaluation protocol are provided in Sec.~\ref{subsec:app:eval_proto} of the appendix). 

\begin{table}[t]
\small
\setlength{\tabcolsep}{1.5pt}
\centering
\begin{tabular}{l|cc|ll}
\toprule
 Method & $R_\omega$ & \# HN & ~~~~Verb$_{H}$ & ~~~~K-400  \\
\midrule
Baseline & $\ind \omega$ & 0 &  ~69.9 & ~55.6 \\
\midrule
w/o calibration &  $\propto B\frac{G_\omega}{S_\omega}$ & 8.7M & ~80.5 \green{\small{(+10.6)}} & ~54.5 \red{\small{(-1.1)}} \\
\rowcolor{aliceblue} w/ ~~calibration &  $\propto \frac{G_\omega}{S_\omega}$, $G_{\omega} \thickapprox  S_{\omega}$ & 0.9M &  ~78.0 \green{\small{(+~~8.1)}}  & ~55.8 \green{\small{(+0.2)}} \\
\bottomrule
\end{tabular}
\caption{\textbf{Importance of the calibration mechanism when training with hard negative captions.}
The model trained without calibration suffers from a drop of performance on Kinetics.
}
\label{tab:naive-hn}
\end{table}

\begin{table}[t]
\setlength{\tabcolsep}{0.1pt}
\centering
\begin{tabular}{l l l}
\toprule
& ~~~~~~~~~~~~\small{w/o calibration} & ~~~~~~~~~~~~~\small{w/ calibration} \\
\vspace{+0.1cm}
\small{$R_{\omega}~~~\propto$} & \multirow{5}{*}{\includegraphics[width=0.4\linewidth]{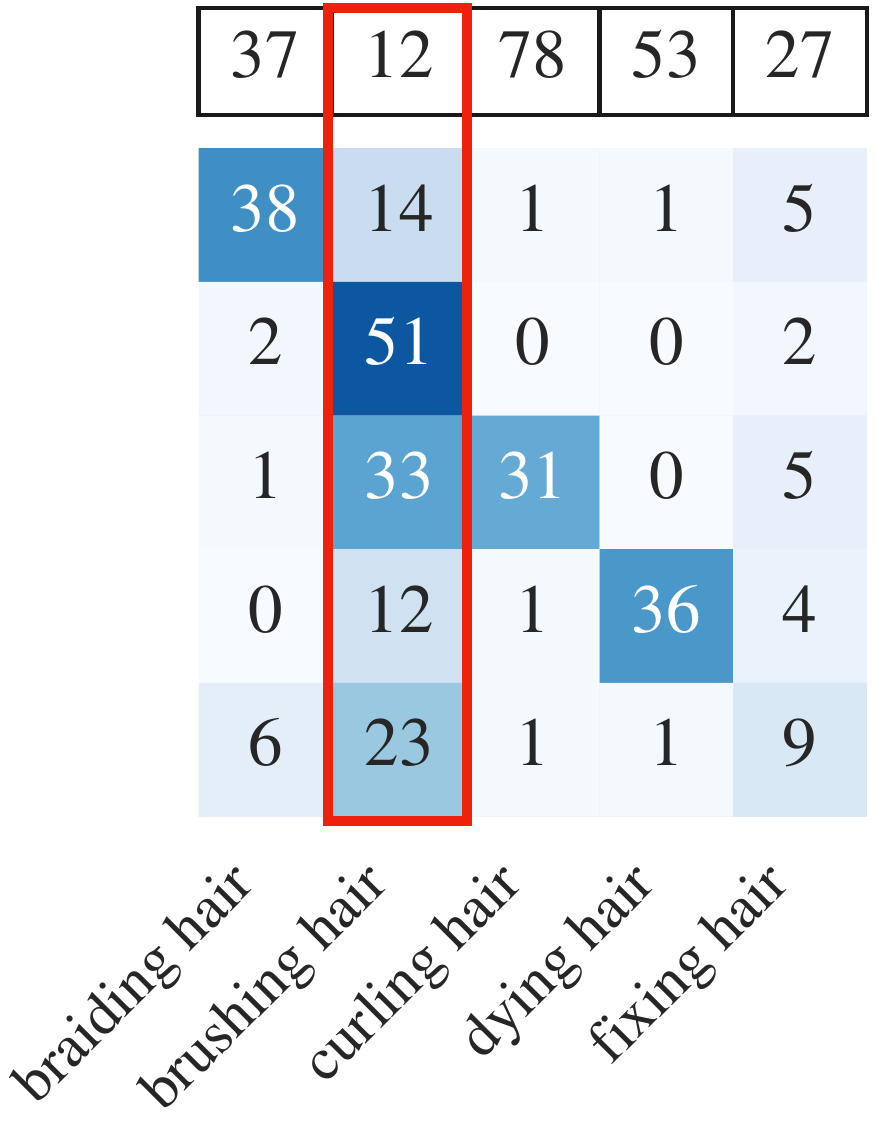}} & \multirow{5}{*}{\includegraphics[width=0.4\linewidth]{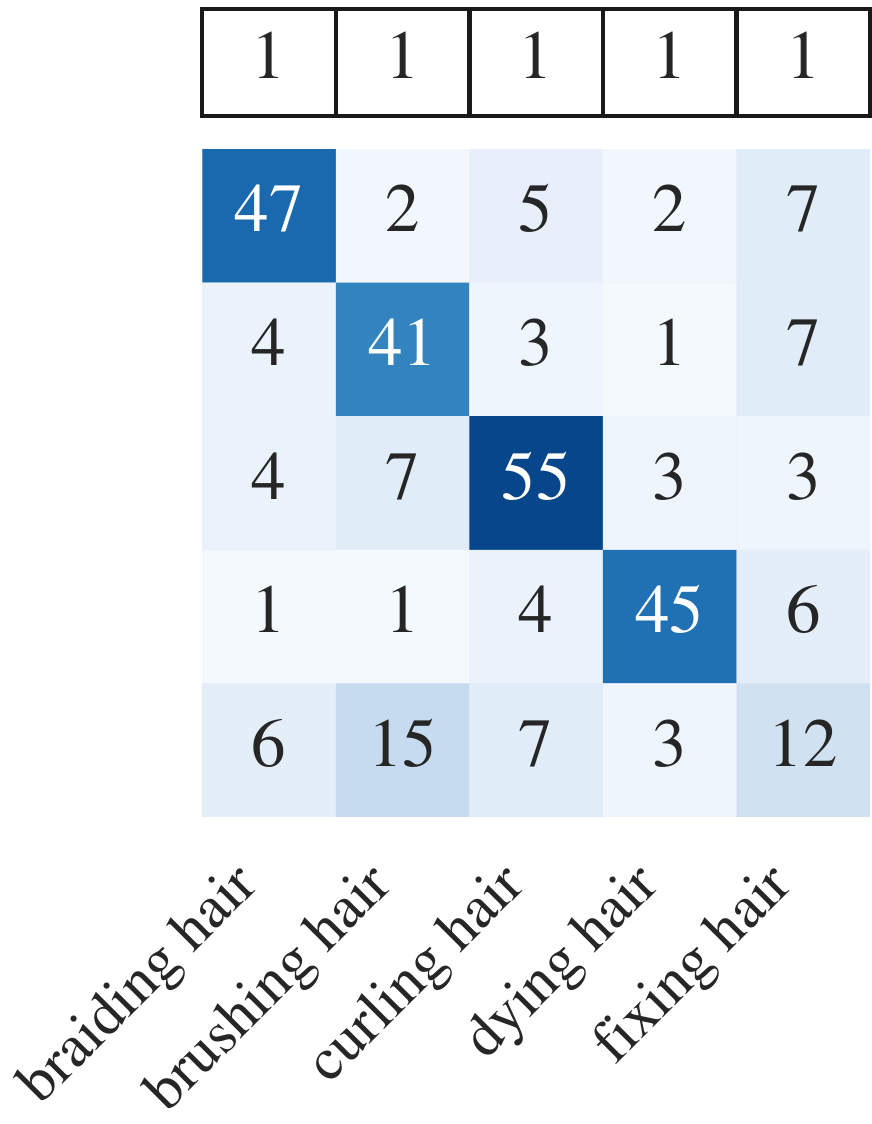}}\\
\vspace{+0.1cm}
\small{braiding hair} & \\
\vspace{+0.1cm}
\small{brushing hair} & \\
\vspace{+0.1cm}
\small{curling hair} & \\
\vspace{+0.1cm}
\small{dying hair} & \\
\vspace{+0.1cm}
\small{fixing hair} & \\
\vspace{+0.8cm}
\end{tabular}
    \caption{
    \textbf{Confusion matrix for the hair classes on Kinetics.}
Without proper calibration, the verb phrase `brushing hair' becomes highly attractive in the video-text feature space.
This deteriorates the performance on all the `hair' related classes.
Our calibration mechanism alleviates this issue by making the ratio $R_\omega$ independent of verb phrases (see details in Sec.~\ref{subsection:neg-methods}). More examples are shown in Sec.~\ref{sec:app:calibration} of the appendix.
    }
    \label{tab:calibration}
    \vspace{-0.4cm}
\end{table}

\subsection{Ablation Study}\label{subsec:ablation}

In this section, we analyze our different design choices.
We report results when transferring the models on two of our benchmarks:
MSR-VTT multi-choice verb split (`Verb$_H$') and Kinetics-400 video classification (`K-400').
We chose these two benchmarks as they have very different properties: the first involves captions, while the second involves action labels. We note that $N^{\text{hard}}=1$ for all ablations unless otherwise specified.

\noindent\textbf{Hard negative captions generation.}
In Tab.~\ref{tab:hn-generation}, we ablate the technique used to obtain additional negative captions: we compare two LLMs (T5~\cite{2020t5} and PaLM~\cite{palm}) and two non LLM-based methods:
(i) `random verb': we replace verbs by random verbs from the UPenn XTag\footnote{\url{https://www.cis.upenn.edu/~xtag/}} verb corpus and
(ii) `antonym verb': we replace verbs with their antonyms, using the NLTK~\cite{bird2009natural} package.
We see in Tab.~\ref{tab:hn-generation} that `random verb' and `antonym verb' already give moderate performance gains on Verb$_H$ compared to the baseline.
However, using LLM-based generations improves the results by a large margin compared to the non LLM-based methods.
This is likely due to the fact that (i) random or antonym replacements often create non semantically or linguistically plausible negative captions; (ii) some verbs do not have antonyms in NLTK (see qualitative examples in Sec.~\ref{sec:app:palm-comparison-neg} of the appendix).
Finally, we see in Tab.~\ref{tab:hn-generation} that T5 generations work very well in our framework too, which demonstrates that our framework is LLM-agnostic and can be extended to other LLMs.
We observe that the best performance is achieved using PaLM, with a substantial gain over the baseline on MSR multi-choice (+8.1\%) and a moderate gain on Kinetics (+0.2\%).

\begin{figure}[t]
\begin{minipage}{0.45\linewidth}
\centering
\small
  \setlength{\tabcolsep}{1.5pt}
   \begin{tabular}{c|cc}
\toprule
PaLM captions: & Verb$_{H}$ & K-400  \\
\midrule
             $\varnothing$ & 69.9 & 55.6 \\
            Positive & 69.3 & 55.4\\
 \rowcolor{aliceblue}        Negative &  \textbf{78.0}  & \textbf{55.8}\\
\bottomrule
\end{tabular}
\end{minipage}\qquad
\begin{minipage}{0.45\linewidth}
\centering
\small
  \setlength{\tabcolsep}{2pt}
\begin{tabular}{c|cc}
\toprule
 Verb isolation: & Verb$_{H}$ & K-400  \\
\midrule
             $\varnothing$ &  69.9 & 55.6 \\
             MiT labels & 69.9 & 57.0 \\
             NLTK~\cite{bird2009natural} & 70.1 & 56.4\\
 \rowcolor{aliceblue}            PaLM~\cite{palm} &  \textbf{70.3}  & \textbf{57.6}\\
\bottomrule
\end{tabular}
\end{minipage}
 \vspace{0.2cm}
\captionof{table}{
(left): \textbf{Generating negative \textit{versus} positive captions with PaLM.}
(right): \textbf{Verb phrase isolation methods.}
}
 \vspace{-0.4cm}
\label{fig:multi_ablat}
\end{figure}

\noindent\textbf{Hard negative captions: the importance of calibration.}
We demonstrate the effect of the calibration mechanism described in Section~\ref{subsection:neg-methods} for training with unpaired captions.
Tab.~\ref{tab:naive-hn} shows the performance of hard negative training with (`w/') \textit{versus} without (`w/o') calibration.
First, we observe that the performance boost on MSR-VTT compared to the baseline is slightly stronger without calibration than with calibration. We believe this is because calibrating the PaLM generations reduces their number. However, we see that training with hard negatives without calibration deteriorates a lot the performance on Kinetics ($-2.0\%$ compared to the baseline).
We hypothesize that this is due to some verb phrases being seen only as repulsive in the video-text feature space, while others are seen equally as attractive and repulsive.
We illustrate this in Tab.~\ref{tab:calibration} by showing the confusion matrix for a subset of the Kinetics classes, along with the ratio $R_\omega$ (defined in Sec.~\ref{subsection:neg-methods}) for each verb phrase.
Intuitively, $R_\omega$ measures the `attraction' (if low) and `repulsion' (if high) of a verb phrase $\omega$.
The confusion matrix in Tab.~\ref{tab:calibration} shows that the verb phrase `brushing hair' becomes an attraction point in the absence of calibration.
Indeed, the number of times the verb phrase `brushing hair' is repulsive versus attractive is low ($R_{\text{brushing hair}} \thickapprox 12$) compared to the other concepts such as for example `curling hair' ($R_{\text{curling hair}} \thickapprox 78$):
we have $R_{\text{brushing hair}} << R_{\text{curling hair}}$.
Hence, predictions for `brushing hair' become dominant.
This actually improves the performance for that class but deteriorates the performance on all the other classes related to `hair'.
We see in Tab.~\ref{tab:calibration} that our calibration mechanism alleviates this effect by making the ratio $R_\omega$ independent of $\omega$ as in regular contrastive learning.
Calibration allows us to improve performance over the baselines on both tasks with a single model. 

\begin{table}[t]
\small
\setlength{\tabcolsep}{1pt}
\centering
\begin{tabular}{lcccll}
\toprule
Method & Hard negatives & Verb phrase  && ~~~Verb$_{H}$ & ~~K-400  \\
\midrule
            Baseline & & && 69.9 & ~55.6 \\
            \midrule
            & \checkmark & && 78.0 \green{\small{(+8.1)}}  & ~55.8 \small{(+0.2)} \\
            & & \checkmark && 70.3 \small{(+0.4)} & ~57.6 \green{\small{(+2.0)}} \\
\rowcolor{aliceblue} \OURS (Ours)  & \checkmark & \checkmark & & 76.3 \green{\small{(+6.4)}} & ~58.5 \green{\small{(+2.9)}} \\
\bottomrule
\end{tabular}
 \vspace{0.2cm}
   \caption{
    \textbf{Combining hard negative and verb phrase loss} 
    achieves 9.2\% relative improvement on MSR-VTT MC (accuracy) and 5.2\% relative improvement on Kinetics (top-1) compared to the baseline.
}
 \vspace{-0.4cm}
    \label{tab:final}
\end{table}

\noindent\textbf{Generating positive \textit{versus} negative captions.}
In Tab.~\ref{fig:multi_ablat} (left), we investigate the impact of generating \textit{positive} captions instead of negatives with PaLM. In this case, positives correspond to sentences where the verb in the original caption is changed to a synonym verb, but the remaining context is unchanged: PaLM therefore acts as a data augmentation generator for text (similar to~\cite{zhao2022lavila, athousandwords}). Details about the positive caption generation implementation are in Sec.~\ref{subsec:app:palm} of the appendix. We observe that using positive captions has a negative impact on the performance in our benchmarks, possibly because with positive captions the model becomes more \textit{invariant} to different verbs.

\noindent\textbf{Verb phrase loss.}
In Tab.~\ref{fig:multi_ablat} (right), we explore two alternatives for verb phrase extraction used in the verb phrase loss:
(i) using human-annotated action labels for clips from the Moments in Time (MiT) dataset (these are available as SMiT data inherits from MiT~\cite{monfortmoments}) and
(ii) using a rule-based method (NLTK~\cite{bird2009natural}) to isolate verbs. 
We observe in Tab.~\ref{fig:multi_ablat} that using PaLM to extract verb phrases from the caption outperforms both, probably because it extracts more fine-grained action information. Qualitative analysis of the verb phrases is shown in Sec.~\ref{sec:app:palm-comparison-vp} of the appendix. 

\noindent\textbf{Combining calibrated hard negatives and verb phrase loss.}
We show in Tab.~\ref{tab:final} the complementarity between our two contributions: the calibrated hard negative training and the verb phrase loss.
The former greatly improves performance on tasks requiring complex language understanding such as MC Verb$_H$.
On the other hand, the verb phrase loss improves transfer to video classification by focusing particularly on the action label in the sentence.
We see in Tab.~\ref{tab:final} that combining both approaches during training results in a \textit{single model} with excellent performance on both MSR-VTT MC and Kinetics zero-shot transfer.
Indeed, compared to the baseline, \OURS pretraining achieves 9.2\% relative improvement on MSR-VTT MC and 5.2\% relative improvement on Kinetics.

\begin{table}[t]
    \setlength{\tabcolsep}{15pt}
    \centering
    \resizebox{0.99\linewidth}{!}{
    \begin{tabular}{lc|cc}
    \toprule
    Method &  $N^{\text{hard}}$ & Verb$_{H}$ & K-400  \\
    \midrule
    VFC (Ours) & 1 &  76.3 & 58.5 \\
       VFC (Ours) & 3 &  77.8 &58.5 \\
  \rowcolor{aliceblue}  VFC (Ours) & 5 & \textbf{78.3} & \textbf{58.5}\\
    \bottomrule
    \end{tabular}
    }
     \vspace{0.2cm}
   \caption{
    \textbf{Maximum number of hard negative captions.} We observe that increasing the maximum number of hard negative captions sampled per video increases the performance on Verb$_{H}$.
    We use $N^{\text{hard}} = 5$ in the remaining of the paper.
}
    \label{tab:num-negs}
\end{table}

\begin{table}[t]
    \setlength{\tabcolsep}{15pt}
    \centering
    \resizebox{0.99\linewidth}{!}{
    \begin{tabular}{lc|cc}
    \toprule
    Method &  Contrastive loss & Verb$_{H}$ & K-400  \\
    \midrule
    Baseline & NCE &  69.9 & 55.6 \\
    Baseline  & HardNeg-NCE &  72.0  & 56.4  \\
    \midrule
    VFC (Ours) & NCE &  78.3 & 58.5 \\
  \rowcolor{aliceblue}  VFC (Ours) & HardNeg-NCE  & \textbf{80.5} & \textbf{58.8} \\
    \bottomrule
    \end{tabular}
    }
     \vspace{0.2cm}
   \caption{
    \textbf{Complementarity with other hard negative mining methods.} We observe that using the HardNeg-NCE loss, instead of standard NCE, gives the highest performance. We use HardNeg-NCE from now on. We note that for VFC we use $N^{\text{hard}}=5$.
}
\vspace{-0.4cm}
    \label{tab:hn-nce}
\end{table}

\noindent\textbf{Number of hard negative captions.} In Tab.~\ref{tab:num-negs}, we experiment with increasing the maximum number of hard negative captions $N^{\text{hard}}$ sampled per video in the batch. We find that setting this to 5 increases the performance on Verb$_{H}$ while maintaining the performance on Kinetics. We use this setting going forward. We note that we do not try larger values as our maximum number of hard negatives per video after calibration is 5.

\noindent\textbf{Complementarity with other hard negative mining methods.}
We investigate whether our VFC framework is complementary to existing approaches for hard negatives with the contrastive learning framework.
Specifically, we reimplement the hard negative noise contrastive multimodal alignment loss from~\cite{rdk+23,Robinson2020ContrastiveLW}, which is denoted as HardNeg-NCE. With this objective, difficult negative pairs (with higher similarity) are emphasised, and easier pairs are ignored. We use $\alpha=1$ and $\beta=0.1$ in the equations from~\cite{rdk+23}. We note that we only adapt $L^{t2v}_i$ and $L^{\text{CHN}}_{i}$ with HardNeg-NCE. Adapting $L^{\text{verb-phrase}}_i$ does not bring further improvements, so we omit this for simplicity.
We observe in Tab.~\ref{tab:hn-nce} that \OURS is complementary to existing hard negative frameworks: using HardNeg-NCE instead of the standard NCE loss achieves the highest performance.
We observe a large boost on Verb$_H$~\cite{park-etal-2022-exposing}, a benchmark that specifically involves hard negatives.
We therefore adopt HardNeg-NCE in the remaining of this paper.

\begin{table}
\small
    \setlength{\tabcolsep}{4pt}
    \centering
        \begin{tabular}{lrc|cc}
            \toprule
  Model &  \# params. && 3k val. & Verb$_{H}$\cite{park-etal-2022-exposing} \\
\midrule
\vspace{+0.05cm}
\textsc{Zero-shot} &&& \\ 
VideoCLIP \cite{xu-etal-2021-videoclip} & -- && 73.9 & - \\
CLIP \cite{Radford2021CLIP} & 151M && 91.1 & 64.1      \\
InternVideo \cite{wang2022internvideo} & $\thickapprox$ 460M && 93.4 & -      \\
\rowcolor{aliceblue} \OURS (Ours)  & 164M && \textbf{95.1} & \textbf{80.5} \\
\midrule
\vspace{+0.05cm}
\textsc{Fine-tuned} && \\ 
ClipBERT~\cite{lei2021less} & -- && 88.2 & - \\
MMT~\cite{gabeur2020mmt} & -- && 92.4 & 71.3 \\
VideoCLIP~\cite{xu-etal-2021-videoclip} & -- && 92.1 & - \\
CLIP-straight~\cite{clip-straight}  & 151M && 94.1 & 65.1 \\
MMT~\cite{gabeur2020mmt}~\footnotesize{(CLIP features)} & -- && 95.0 & 71.4 \\
C4CL-mP~\cite{park-etal-2022-exposing} & 151M && \textbf{96.2} & 73.7   \\
\rowcolor{aliceblue} \OURS (Ours)  & 164M && \textbf{96.2} & \textbf{85.2} \\
\bottomrule
        \end{tabular}
 \vspace{0.2cm}
    \caption{
    \textbf{Multi-choice MSR-VTT.}
We report accuracy on the 3k val and on the verb-focused Verb$_H$~\cite{park-etal-2022-exposing} splits.
While \OURS improves the performance on both splits in a zero-shot setting, the gap with previous works is especially important on Verb$_H$~\cite{park-etal-2022-exposing}, a split measuring verb understanding.
When available, we add model parameter counts.
    }
    \label{tab:main-results-MSR-VTT}
    \vspace{-0.2cm}
\end{table}

\subsection{Comparisons to the State of the Art}\label{subsec:results}

We compare our \OURS features to the state of the art on a diverse set of tasks requiring verb understanding.
Note that we use the \textit{same model} across different tasks, which is non-trivial in itself as the tasks cover a wide range of domains and evaluation protocols. 

\noindent\textbf{MSR-VTT MC results.} 
We see in Tab.~\ref{tab:main-results-MSR-VTT} that our verb-focused pretraining transfers well to the MSR-VTT multi-choice task, especially on the hard verb split (curated to assess exactly the task we are trying to solve).
We even outperform concurrent InternVideo~\cite{wang2022internvideo} while using a significantly smaller setting both in terms of architecture (InternVideo uses 2.8$\times$ more parameters and 12.4$\times$ more flops) and pretraining dataset size (they use 24$\times$ more data).
We also note that our method does not degrade performance on other standard object-based tasks, such as text-to-video retrieval on MSR-VTT (results compared to the state of the art are shown in Sec.~\ref{subsec:app:retrieval} of the appendix).

\begin{table}[t]
\small
    \setlength{\tabcolsep}{4pt}
    \centering
        \begin{tabular}{l|cccc|ccc}
            \toprule
                       &  \multicolumn{4}{c|}{}    &  \multicolumn{3}{c}{ATP$_{hard}$ \cite{buch2022revisiting}}  \\
Model  & all  & D & T & C  &  all  & T & C \\
\midrule
\vspace{+0.05cm}
        \textsc{Zero-shot} \\
        CLIP \cite{Radford2021CLIP}  & 43.9 & 57.0 & 38.1 & 43.6 &23.0 &21.8 & 23.8\\
          \rowcolor{aliceblue} \OURS (Ours) &  \textbf{51.5} & \textbf{64.1} & \textbf{45.4} & \textbf{51.6} & \textbf{31.4} & \textbf{30.0} & \textbf{32.2}\\
        \midrule 
        \vspace{+0.05cm}
        \textsc{Fine-tuned} \\ 
         \gray{HGA$\ddagger$} \cite{jiang2020reasoning} & \gray{49.7} & \gray{59.3} & \gray{50.7} & \gray{46.3} & \gray{44.1}  & \gray{45.3} & \gray{43.3}   \\
         ATP \cite{buch2022revisiting}  &   49.2 & 58.9 & 46.7 & 48.3 & 20.8 &  22.6 & 19.6 \\
         Temp[ATP] \cite{buch2022revisiting} & 51.5 & 65.0 & 49.3 & 48.6 & 37.6 & 36.5 & 38.4    \\
         TAATP$\dagger$ \cite{xiao2022video}&  54.3 & 66.8 & 50.2 & 53.1  & - & - & - \\
         VGT \cite{xiao2022video} &55.0 & 64.1 & \textbf{55.1} & 52.3 & - & - & - \\
          \rowcolor{aliceblue}  \OURS (Ours) &  \textbf{58.6} & \textbf{72.8} & 53.3 & \textbf{57.6} & \textbf{39.3} & \textbf{38.3} & \textbf{39.9}\\
            \bottomrule
        \end{tabular}
 \vspace{0.2cm}
    \caption{
    \textbf{NEXT-QA video question answering.}
We report accuracy.
We consider either `all' questions or only causal (`C'), temporal (`T') or  descriptive (`D') questions.
We also use ATP$_{hard}$ split~\cite{buch2022revisiting}.
\OURS improves performance for both zero-shot and fine-tuning.
$\dagger$Temp[ATP]+ATP.
$\ddagger$ Uses additional motion features.}
    \label{tab:main-results-next-qa}
    \vspace{-0.3cm}
\end{table}

\noindent\textbf{NEXT-QA results.}
We show in Tab.~\ref{tab:main-results-next-qa} that our verb-focused pretraining gives a significant boost in both the standard and ATP$_{hard}$ setting introduced by~\cite{buch2022revisiting}.
To the best of our knowledge, we are the first work to report zero-shot results for NEXT-QA and our zero-shot numbers improve upon some previously published fine-tuning numbers.
Finally, although HGA~\cite{jiang2020reasoning} performs worse than ours on the standard setting, it achieves a high accuracy of 44.1 on ATP$_{hard}$.
Their high performance on ATP$_{hard}$ can be explained by the use of additional motion features, aiding in answering hard dynamics questions, as noted by~\cite{buch2022revisiting}.
The addition of extra motion features on the video side can be complementary to our verb-focused pretraining approach.

\noindent\textbf{Zero-shot Kinetics-400 results.}
In Tab.~\ref{tab:main-results-kinetics} we see that our verb-focused features transfer very well to Kinetics video classification benchmark in a zero-shot setting, achieving state-of-the-art results. 
We achieve better results than Flamingo models~\cite{flamingo} while using a significantly smaller model: relative improvement of 20\% over Flamingo-80B model while using 489 $\times$ less parameters.

\begin{table}[t]
\small
    \setlength{\tabcolsep}{5.5pt}
    \centering
        \begin{tabular}{lrr|ccc}
            \toprule
            Model & \# param.  && top-1 & top-5 & average \\
            \midrule 
        \textsc{Val-Set} \\
         CLIP \cite{Radford2021CLIP} & ~~151M && 48.9  & 75.8  & 62.4 \\
        ActionCLIP~\cite{actionclip} & $\thickapprox$ 164M && 56.4 & - & - \\
        \rowcolor{aliceblue} \OURS (Ours) & ~~164M && \textbf{59.4} &\textbf{85.3} &\textbf{72.4} \\  
        \midrule
        \textsc{Test-Set} \\
        Flamingo-3B~\cite{flamingo} & ~~3B &&45.2 & 66.8  & 56.0\\
        Flamingo-80B~\cite{flamingo} & ~~80B &&49.1 & 71.5 & 60.3 \\
        Flamingo-9B~\cite{flamingo} & ~~9B &&49.7 & 71.5 & 60.6 \\
        CLIP \cite{Radford2021CLIP} & ~~151M && 47.9  & 75.1  & 61.5 \\
        \rowcolor{aliceblue} \OURS (Ours) & ~~164M && \textbf{58.8} &\textbf{84.5} &\textbf{71.7} \\  
        
        \bottomrule
        \end{tabular}
         \vspace{0.2cm}
    \caption{
    \textbf{Zero-shot transfer to Kinetics-400.}
    We report top-1 accuracy, top-5 accuracy, and their average on the validation and test set, as well as the parameter counts of the different models.
   }
    \label{tab:main-results-kinetics}
    \vspace{-0.2cm}
\end{table}

\begin{table}[t]
\small
    \setlength{\tabcolsep}{22pt}
    \centering
        \begin{tabular}{l|cc}
            \toprule
  Model   & top-1 & top-5 \\
\midrule
\textsc{Zero-shot} &\\ 
CLIP \cite{Radford2021CLIP}  & 59.7  & 83.9  \\
\rowcolor{aliceblue} VFC (Ours)  & \textbf{70.2} & \textbf{92.5}  \\
\midrule
\textsc{Fine-tuned} & \\ 
ER-ZSAR~\cite{Chen2021ElaborativeRF}  & 42.1 & 73.1 \\
X-CLIP~\cite{XCLIP}  & 65.2 & 86.1 \\
X-Florence~\cite{XCLIP} &  68.8 & 88.4 \\
\bottomrule
        \end{tabular}
    \vspace{+0.2cm}
    \caption{
    \textbf{Zero-shot transfer to Kinetics-600.} We report average top-1 and top-5 accuracies over three random 160-class splits, covering classes not in Kinetics-400 but within Kinetics-600. While~\cite{Chen2021ElaborativeRF, XCLIP} fine-tune on Kinetics-400, we surpass their performance in zero-shot.}
    \vspace{-0.2cm}
    \label{tab:kinetics-600}
\end{table}

\noindent\textbf{Zero-shot Kinetics-600 results.} We evaluate our model on Kinetics-600 in Tab.~\ref{tab:kinetics-600} and follow the protocol in~\cite{XCLIP, Chen2021ElaborativeRF}. Specifically, the subset of categories which are outside Kinetics-400, but within Kinetics-600 are used for evaluation. The evaluation is then run on a random sample of 160 categories from this subset. The final performance is averaged over three iterations. We observe that by evaluating our model in a zero-shot setting, we surpass the performance of works~\cite{Chen2021ElaborativeRF, XCLIP} which fine-tune on Kinetics-400.

\begin{table}[t]
\small
\setlength{\tabcolsep}{14pt}
\centering
\begin{tabular}{l|ll}
\toprule
Method & ~~~~~~~~~all & ~~Kinetics-verb  \\
\midrule
Baseline & ~~~55.6 & ~~~52.1 \\
  \rowcolor{aliceblue}       \OURS (Ours)   & ~~~\textbf{58.8} \green{\small{(+3.2)}} & ~~~\textbf{57.1} \green{\small{(+5.0)}} \\
\bottomrule
\end{tabular}
\vspace{0.2cm}
    \caption{
    \textbf{Zero-shot Kinetics-verb.}
    We report accuracy performance on our newly proposed Kinetics-verb split (from test split).}
    \vspace{-0.2cm}
    \label{tab:kinetics_verbs}
\end{table}
\noindent\textbf{Kinetics-verb.}
To further analyse the \OURS framework's effect on action classification, we introduce the Kinetics-verb split.
We isolate classes from the Kinetics-400 dataset that share a common noun with another class, but have a different verb (and therefore action).
For example, distinguising between `braiding hair', `brushing hair' and `curling hair' requires the model to focus on verb understanding as predictions cannot be inferred from the simple presence of hair in the frame.
We use this rule to create a subset of 97 classes from the Kinetics-400 test set (see Sec.~\ref{subsec:app:kinetics-verb} in the appendix) called `Kinetics-verb'.
We show in Tab.~\ref{tab:kinetics_verbs} that our \OURS improves substantially over the baseline (+5\%) on this split.

\noindent\textbf{Assessing verb understanding on SVO-probes.}
In Tab.~\ref{tab:svo-probes}, we see that our \OURS framework improves the performance on SVO-probes compared to the baseline (particularly in the verb setting), and outperforms prior work~\cite{hendricks2021probing} with 21.7\% relative improvement in the verb setting.

\begin{table}[t]
\small
\setlength{\tabcolsep}{18pt}
\centering
\begin{tabular}{l|cc}
\toprule
Model & AP & AP$_\text{verb}$  \\
\midrule
            CLIP~\cite{Radford2021CLIP} & 48.3 & 52.3 \\
            No-MRM-MMT~\cite{hendricks2021probing}$\dagger$ & 51.5 & 53.1 \\
            Baseline (Ours) & 60.2 & 61.9 \\
  \rowcolor{aliceblue}      \OURS (Ours)   & \textbf{61.8}  & \textbf{64.6}\\
\bottomrule
\end{tabular}
\vspace{0.2cm}
    \caption{
    \textbf{Verb understanding on SVO-probes~\cite{hendricks2021probing}.}
    We report Average Precision (AP) on the entire dataset and on the verb-specialized setting. $\dagger$ Scores provided by authors and used to calculate AP. 
    }
    \vspace{-0.2cm}
    \label{tab:svo-probes}
\end{table}

\section{Conclusion}\label{sec:conclusion}

Video-language models based on CLIP have been shown to have limited verb understanding, relying extensively on nouns.
We attempt to alleviate this problem with two technical contributions on the contrastive learning framework:
first, we leverage LLMs to automatically generate hard negative captions focused on verbs;
second, we introduce a verb phrase alignment loss.
We validate our verb-focused pretraining by showing improved performance on a suite of benchmarks, chosen in particular to assess verb understanding.
Our framework is general and could be employed for other video-language tasks, and further readily scales with the rapid progress in language modelling.
\\

{
\noindent\textbf{Acknowledgements.}~We would like to thank Ahmet Iscen, Anurag Arnab, Paul Hongsuck Seo, Antoine Yang,  Shyamal Buch, Alex Salcianu for their precious help and discussions. We also thank Sagar Vaze for his invaluable support.
\par
}
\clearpage

{\small
\bibliographystyle{ieee_fullname}
\bibliography{references}
}
\clearpage

\def\OURS{VFC\xspace}
\vspace{-1cm}
\maketitle
\vspace{-1cm}
\appendix

\part{Appendix} 
{\hypersetup{linkcolor=black} \parttoc}

\clearpage
\ificcvfinal\thispagestyle{empty}\fi

\renewcommand{\thefigure}{A.\arabic{figure}}
\setcounter{figure}{0} 
\renewcommand{\thetable}{A.\arabic{table}}
\setcounter{table}{0}

This appendix to the main paper provides additional quantitative 
(Sec.~\ref{sec:app:exp})
and qualitative results
(Sec.~\ref{sec:app:qualitative}),
and further details on baselines and implementation
(Sec.~\ref{sec:app:details}).

\section{Quantitative results}\label{sec:app:exp}

In this section, we present results comparing standard versus Verb-Focused Constrastive (VFC) learning for all benchmarks (Sec.~\ref{subsec:app:base_vs_vfc}), comparison to state-of-the-art methods for MSR-VTT retrieval (Sec.~\ref{subsec:app:retrieval}), and additional ablations (Sec.~\ref{subsec:app:ablations}).

\subsection{Standard vs.~Verb-Focus Contrastive (VFC) learning for all benchmarks}\label{subsec:app:base_vs_vfc}
We see in Tab.~\ref{tab:baseline-versus-ours} that our VFC learning performs better than standard contrastive learning (Baseline) for all verb-focused benchmarks on both zero-shot and fine-tuned settings while maintaining performance on more noun-focused benchmarks, such as MSR-VTT random MC. We observe that using the HardNeg-NCE loss, instead of standard NCE, further improves performance for all benchmarks on both zero-shot and fine-tuned settings.

\begin{table}
    \setlength{\tabcolsep}{2pt}
    \centering
    \resizebox{\linewidth}{!}{
        \begin{tabular}{lc|cc|cc|cc|cc}
            \toprule
         &    &\multicolumn{2}{c|}{\textbf{MSR-VTT}} &    \multicolumn{2}{c|}{\textbf{K-400}} &   \multicolumn{2}{c|}{\textbf{NEXT-QA}} &   \multicolumn{2}{c}{\textbf{SVO}} \\
        &    & 3k val. & Verb$_H$ & all & verb & all & ATP$_{hard}$ & all & verb \\
  Method & loss    & MC & MC & top-1 & top-1 & MC & MC & AP & AP \\
\midrule
\textsc{Zero-shot} &\\ 
Baseline & NCE & 94.9 & 69.9  & 55.6 & 52.1 & 48.6 & 28.9 & 60.2 & 61.9 \\
\rowcolor{aliceblue} VFC & NCE & 94.9 & 78.3  & 58.5 & 56.7 & 51.0 & 31.3 & 61.5 & 63.9 \\
\rowcolor{aliceblue} VFC & HardNeg-NCE&  \textbf{95.1} & \textbf{80.5}  & \textbf{58.8} & \textbf{57.1} & \textbf{51.5} & \textbf{31.4} & \textbf{61.8} & \textbf{64.6} \\
\midrule
\midrule
\textsc{Fined-tuned} & \\ 
Baseline  & NCE & \textbf{96.8} & 73.8  & - & - & 57.3 & 37.8 & - & - \\
\rowcolor{aliceblue}  VFC & NCE & 96.2 & 84.8  & - & - & 58.4 & 38.3 & - & - \\
\rowcolor{aliceblue}  VFC & HardNeg-NCE & 96.2 & \textbf{85.2}  & - & - & \textbf{58.6} & \textbf{39.3} & - & - \\
\bottomrule
        \end{tabular}
    }
    \caption{
        \textbf{Standard vs.~Verb-Focus Contrastive learning for all benchmarks.} We report MSR-VTT random (3k val.) and Verb$_H$~\cite{park-etal-2022-exposing} multiple-choice accuracies, Kinetics-400 and Kinetics-verb top-1 accuracies, NEXT-QA and ATP$_{hard}$~\cite{buch2022revisiting} multiple-choice accuracies, and SVO-probes entire dataset and verb-focused Average Precision. We observe that our VFC learning performs better than standard contrastive learning (Baseline) for all verb-focused benchmarks on both zero-shot and fine-tuned settings, while maintaining performance on more noun-focused benchmarks, such as MSR-VTT random MC. We observe that using the HardNeg-NCE loss, instead of standard NCE, further improves performance for all benchmarks on both zero-shot and fine-tuned settings. }
    \label{tab:baseline-versus-ours}
\end{table}

\subsection{MSR-VTT retrieval}\label{subsec:app:retrieval}
We see in Tab.~\ref{tab:retrieval-MSR-VTT} that while our verb-focused pretraining drastically improves performance on verb-focused benchmarks -- such as Verb$_H$ split~\cite{park-etal-2022-exposing} MSR-VTT (see main paper Tab.~\ref{tab:main-results-MSR-VTT}) -- it maintains performance on noun-focused benchmarks such as MSR-VTT retrieval T2V (1k split) in a zero-shot setting. We perform comparably to InternVideo~\cite{wang2022internvideo} in a zero-shot setting, while using a significantly smaller setting both in terms of architecture (Intern-Video uses 2.8× more parameters and 12.4× more flops) and pretraining dataset size (they use 24× more data). In a fine-tuned setting, InternVideo surpasses VFC's performance. This is expected given our model parameters and flops are significantly smaller -- see number of parameters in Tab.~\ref{tab:retrieval-MSR-VTT}.

\begin{table}
    \setlength{\tabcolsep}{12pt}
    \centering
    \resizebox{0.99\linewidth}{!}{
        \begin{tabular}{l|cc}
            \toprule
                         \multicolumn{1}{c|}{} & & 1K val. \\
  Model   & \# params. & T$\rightarrow$V  R@1 \\
\midrule
\textsc{Zero-shot} &\\ 
VideoCLIP \cite{xu-etal-2021-videoclip} & -- & 10.4 \\
CLIP \cite{Radford2021CLIP}  & 151M  & 30.6  \\
InternVideo \cite{wang2022internvideo}$\ddagger$   & $\thickapprox$ 460M  & \textbf{40.7}  \\
\midrule
\rowcolor{aliceblue} VFC (Ours)  &164M  & 40.3 \\
\midrule
\midrule
\textsc{Fined-tuned} & \\ 
ClipBERT \cite{lei2021less}  & -- & 22.0 \\
MMT~\cite{gabeur2020mmt} & --  & 26.6 \\
VideoCLIP \cite{xu-etal-2021-videoclip}  & -- & 30.9 \\
CLIP-straight~\cite{clip-straight} &  151M & 31.2 \\
MMT (CLIP features) ~\cite{gabeur2020mmt}  & -- & 34.0 \\
C4CL-mP~\cite{park-etal-2022-exposing}  & 151M  & 43.1 \\
CLIP2Video~\cite{park-etal-2022-exposing}  &--  & 45.6 \\
InternVideo \cite{wang2022internvideo}$\ddagger$   & $\thickapprox$ 460M  & \textbf{55.2}  \\
  \midrule
\rowcolor{aliceblue}  VFC (Ours) & 164M & 44.5 \\
\bottomrule
        \end{tabular}
    }
    \vspace{0.2cm}
    \caption{
    \textbf{Results on MSR-VTT retrieval.}
We report T2V retrieval on the 1k split. While our VFC framework drastically improves performance on verb-focused benchmarks, including Verb$_H$ split~\cite{park-etal-2022-exposing} (see main paper Tab.~\ref{tab:main-results-MSR-VTT}), it maintains performance on noun-focused benchmarks such as the retrieval 1k split in the zero-shot setting. In the fine-tuned setting, InternVideo surpasses VFC's performance. $\ddagger$ InternVideo is concurrent unpublished work with a larger model (2.8× more parameters and 12.4× more flops), and has a larger pretraining dataset size (they use 24× more data). }
    \label{tab:retrieval-MSR-VTT}
\end{table}

\subsection{Additional ablations}\label{subsec:app:ablations}

Here, we present ablation results for video mining (Sec.~\ref{subsec:video-mining}), PaLM prompting (Sec.~\ref{subsec:input-output}), the verb phrase loss (Sec.~\ref{subsec:verb-phrase-ablation}), fine-tuning strategy (Sec.~\ref{subsec:freezing}) and calibration (Sec.~\ref{subsec:calibration-ablation}). We note that all ablations are performed with the standard NCE loss (not HardNeg-NCE).

\subsubsection{Video mining}\label{subsec:video-mining}

\begin{figure*}[ht!]
    \centering
    \includegraphics[width=\textwidth]{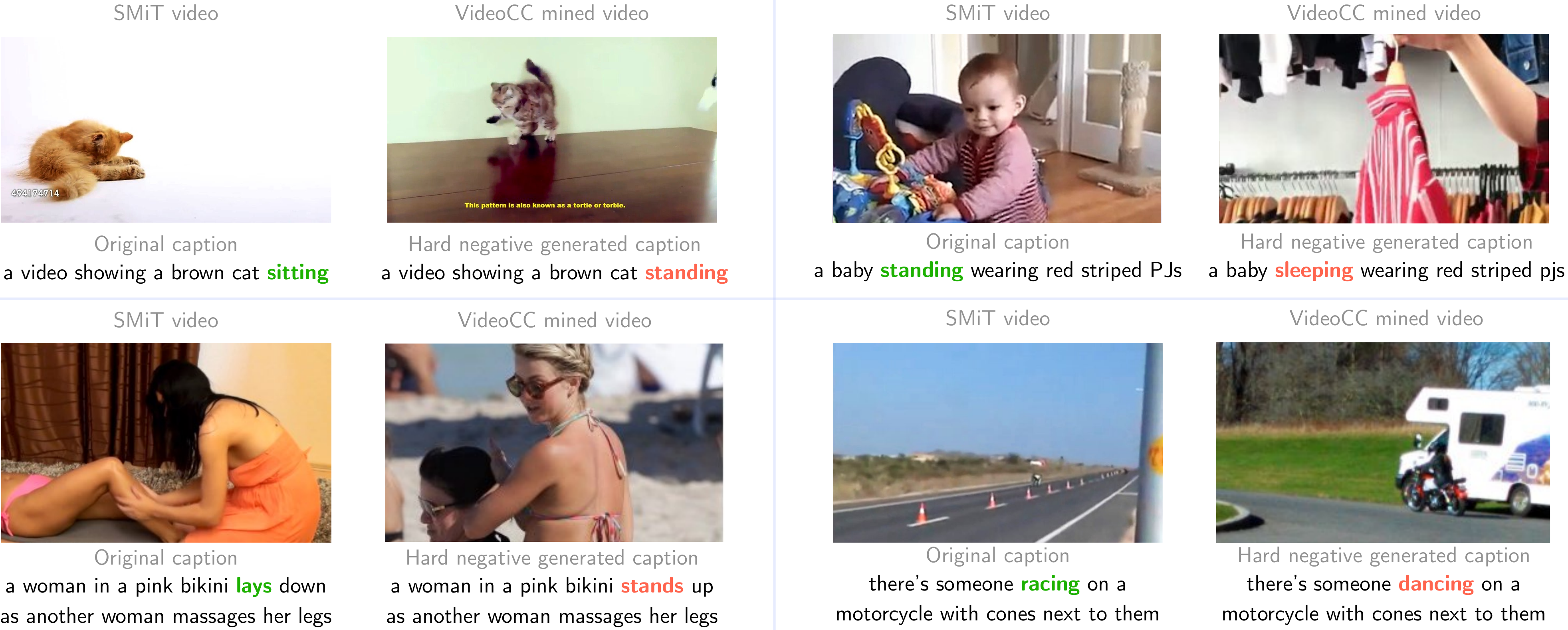}
    \caption{\textbf{Video mining:} We show examples of mined matching videos for generated hard negative captions. For ease of visualisation, we show a single frame per video. In some cases, as the top left corner, the mined video from VideoCC closely matches the hard negative caption. However, often, the new video-text pairs are noisy. For example, in the top right corner, the mined video contains a `red striped shirt' but no `baby sleeping'. In the bottom left example, there is a `woman in a pink bikini standing up' but no `woman massaging her legs'. Finally, in the bottom right example, although the video contains a `motorcycle', the person is not `dancing' and there are no `cones next to them'. }
    \label{fig:video-mining}

\end{figure*}

\begin{table}
    \setlength{\tabcolsep}{8pt}
    \centering
    \resizebox{0.99\linewidth}{!}{
    \begin{tabular}{lc|lll}
    \toprule
    Method & \# pairs & ~~~~Verb$_{H}$ & ~~~~K-400  & ~~~~SMiT \\
    \midrule
    Baseline & 481K &  69.9 & 55.6 & 78.3 \\
    HN & 481K & 78.0 \green{\small{(+8.1)}}& 55.8 \gray{\small{(+0.2)}} & 78.6 {\gray{\small(+0.3)}}\\
    HN+VM & 1.22M & 78.7 \green{\small{(+8.8)}}& 51.8 \red{\small{(-3.8)}}& 75.0 \red{\small{(-3.3)}}\\
    \bottomrule
    \end{tabular}
    }
        \vspace{0.2cm}
    \caption{
    \textbf{Video Mining.} We report multi-choice accuracy on Verb$_H$~\cite{park-etal-2022-exposing}, Kinetics-400 top-1 accuracy and V2T R@1 on Spoken Moment in Time (validation set of our pretraining SMiT data). We observe that although our Video Mining (VM) approach improves performance on Verb$_H$, it causes a drop in performance on Kinetics and SMiT, which highlights that the additional video-text pairs are noisy. For experiments including hard negatives, we note that one hard negative is sampled for each video here.}
    \label{tab:video-mining}
\end{table}

An alternative to our proposed calibration strategy to avoid imbalances due to the addition of negative captions would be to avoid training with unpaired data at all, by mining a matching video $V^{\text{hard}}_{i_k}$ for each generated caption $T^{\text{hard}}_{i_k}$. 
We attempt this via CLIP-based text-to-video retrieval in a large video database. We next explain our pipeline in more detail.

Firstly, we generate hard negative verb captions with PaLM as explained in Sec.~\ref{sec:vfc} of the main paper. For each hard negative caption, we then perform text-to-image retrieval to find a matching video in the VideoCC~\cite{nagrani2022learning} database. Specifically, we calculate the cosine similarity between the hard negative caption CLIP text embedding and the average of the video frames' CLIP image embedding, for all videos in the database. We then keep the video with closest similarity to the hard negative caption to form a new video-text pair.
Finally, we apply a similarity threshold to keep only the best matching video-text pairs and add these to our training set. In practice, we experiment with different thresholds and find a value of 0.28 to work best, adding a total of 738K new video-text pairs to training (SMiT training set size is 481K). Note that we also experiment with text-to-text retrieval: in this case, we calculate the similarity between each hard negative caption and all VideoCC captions, and subsequently use the video corresponding to the closest VideoCC caption to form a new pair. However, we find this performs worse.  

We observe in Tab.~\ref{tab:video-mining} that our additional video-text pairs are noisy. In fact, although this approach improves performance on Verb$_H$, it causes a large drop in performance on Kinetics and SMiT (validation set of our pretraining data). Finding a video matching a specific, detailed and long caption is challenging (see qualitative examples in Fig.~\ref{fig:video-mining}). A video matching the caption may not exist in the VideoCC corpus and even if it did, for this method to be successful, the mined video must match the generated caption on the verbs (and CLIP is biased towards images and objects only, which is exactly the problem we are trying to solve).

\subsubsection{Giving input-output example pairs to PaLM}\label{subsec:input-output}
To generate hard verb negative captions with PaLM, we also add four input-output pair examples to the prompt (see full prompt in Sec.~\ref{subsec:app:palm}) to increase the quality of the generated hard negatives. We observe in Tab.~\ref{tab:input-output} that the input-output pairs improve the performance on Verb$_H$ and Kinetics-400.

\begin{table}
    \setlength{\tabcolsep}{22pt}
    \centering
    \resizebox{0.99\linewidth}{!}{
    \begin{tabular}{c|cc}
    \toprule
    input-output pairs & Verb$_{H}$ & K-400  \\
    \midrule
      &  77.5 & 54.6 \\
    \checkmark &  78.0 & 55.8 \\
    \bottomrule
    \end{tabular}
    }
     \vspace{0.2cm}
    \caption{
    \textbf{Inclusion of input-output pairs in PaLM prompt.} We report multi-choice accuracy on Verb$_H$~\cite{park-etal-2022-exposing} and Kinetics-400 top-1 accuracy. We observe that including input-output pairs in the PaLM prompt for generating hard negative captions increases the performance on both benchmarks. We note that one hard negative is sampled for each video here.}
    \label{tab:input-output}
 
\end{table}

\subsubsection{Verb phrase loss}\label{subsec:verb-phrase-ablation}
We see in Tab.~\ref{tab:verb-phrase-v2t} that using only the video-to-text component of the verb phrase loss allows us to maintain performance on noun-focused benchmarks such as MSR-VTT retrieval, while also giving a performance boost on verb focused benchmarks Verb$_H$ and K-400. 

\begin{table}
    \setlength{\tabcolsep}{4pt}
    \centering
    \resizebox{0.99\linewidth}{!}{
        \begin{tabular}{lcc|ll|l}
            \toprule
             & & &  \multicolumn{2}{c|}{\textbf{MSR-VTT}} &  ~~~~\textbf{K-400} \\
             &\multicolumn{2}{c|}{Verb phrase loss}  & ~~~1k val. &  ~Verb$_H$ &  ~~~~~~~all \\
  Method   &  T$\rightarrow$V  & V$\rightarrow$T  & T$\rightarrow$V  R@1 & MC acc & ~~~~Top-1 \\
\midrule
 Baseline &   &  & 40.8 & 69.9 & 55.6 \\
 VFC (Ours) & \checkmark & \checkmark & 38.8 \red{\small{(-2.0)}} & 77.0 \green{\small{(+7.1)}} & 58.8 \green{\small{(+3.2)}}\\
 VFC (Ours) &  & \checkmark & 40.1 \gray{\small{(-0.7)}} & 76.3 \green{\small{(+6.4)}} & 58.5 \green{\small{(+2.9)}} \\
\bottomrule
        \end{tabular}
    }
     \vspace{0.2cm}
    \caption{
    \textbf{Verb phrase loss.}
We report MSR-VTT T2V retrieval on the 1k split, multi-choice accuracy on Verb$_H$~\cite{park-etal-2022-exposing} and Kinetics-400 top-1 accuracy. We observe that using only the video-to-text component of the verb phrase loss allows us to maintain performance on noun-focused benchmarks such as MSR-VTT retrieval, while also giving a performance boost on Verb$_H$ and K-400. For experiments including hard negatives, we note that one hard negative is sampled for each video here.}
    \label{tab:verb-phrase-v2t}
\end{table}

\subsubsection{Fine-tuning image and text towers}\label{subsec:freezing}
We experiment with different fine-tuning strategies: (i) fine-tuning both image and text towers, (ii) freezing the image CLIP backbone only (here, the sequence Transformer seqTrans and text tower are trained -- see Sec.~\ref{subsec:app:clip4clip} for more details on the CLIP4CLIP architecture), (iii) freezing the text tower only (here, seqTrans and image CLIP backbone are trained), (iv) freezing both image and text towers (here, only seqTrans is trained). We see in Tab.~\ref{tab:finetuning-image-text} that fine-tuning both image and text towers works best. We do not include setting (iv) as it performs very poorly.

\begin{table}
    \setlength{\tabcolsep}{6pt}
    \centering
    \resizebox{0.99\linewidth}{!}{
    \begin{tabular}{lcc|ll}
    \toprule
    Method &  \SnowflakeChevron text & \SnowflakeChevron image & ~~~Verb$_{H}$ & ~~~~K-400  \\
    \midrule
    Baseline & & & 69.9 & 55.6 \\
    \midrule
    VFC (Ours) &  & &  76.3 & 58.5 \\
    VFC (Ours)& \checkmark & &  72.0 \red{\small{(-4.3)}} & 54.8 \red{\small{(-3.7)}} \\
    VFC (Ours) & & \checkmark  & 75.1 \red{\small{(-1.2)}} & 55.1 \red{\small{(-3.4)}} \\
    \bottomrule
    \end{tabular}
    }
     \vspace{0.2cm}
    \caption{
    \textbf{Fine-tuning image and text towers.} We report multi-choice accuracy on Verb$_H$~\cite{park-etal-2022-exposing} and Kinetics-400 top-1 accuracy. \SnowflakeChevron corresponds to freezing the image or text tower. We observe that fine-tuning both image and text towers works best. For experiments including hard negatives, we note that one hard negative is sampled for each video here.}
    \label{tab:finetuning-image-text}
\end{table}

\subsubsection{Calibration}\label{subsec:calibration-ablation}
As explained in Sec.~\ref{sec:vfc} of the main paper, our calibration strategy is composed of two steps: (1) ignoring hard negative captions from the other elements of the batch (denoted as `reducing $B$ effect', where $B$ is the batch size); (2) filtering the generated PaLM captions to have equal number of concept occurences in positive and negative pairs (denoted as $G_\omega \thickapprox S_\omega$). We show the effect of each of these steps in Tab.~\ref{tab:calibration-steps}. We observe that by combining both steps, we avoid a drop in performance on Kinetics-400, while maintaining a large performance improvement on Verb$_H$.

\begin{table}
    \setlength{\tabcolsep}{6pt}
    \centering
    \resizebox{0.99\linewidth}{!}{
    \begin{tabular}{lcc|cc}
    \toprule
    Method &  reducing $B$ effect& $G_\omega \thickapprox S_\omega$ & Verb$_{H}$ & K-400  \\
    \midrule
    Baseline  & & & 69.9 & 55.6 \\
    HN & & & 80.5 & 54.5 \\
    HN & \checkmark & &  79.4 & 55.4 \\
    HN&  & \checkmark&  78.7 & 55.2 \\
    HN &\checkmark & \checkmark  & 78.0 & 55.8 \\
    \bottomrule
    \end{tabular}
    }
     \vspace{0.2cm}
    \caption{
    \textbf{Calibration strategy.} We report multi-choice accuracy on Verb$_H$~\cite{park-etal-2022-exposing} and Kinetics-400 top-1 accuracy. We observe that by combining both calibration steps, we avoid a drop in performance on Kinetics-400, while maintaining a large performance improvement on Verb$_H$. For experiments including hard negatives, we note that one hard negative is sampled for each video here.}
    \label{tab:calibration-steps}
\end{table}

\vspace{+1cm}
\section{Qualitative results}\label{sec:app:qualitative}

In this section, we present qualitative results on MSR-VTT (Sec.~\ref{sec:app:msr-vtt}) and NEXT-QA (Sec.~\ref{sec:app:next-qa}), further analysis of calibration on Kinetics-verb (Sec.~\ref{sec:app:calibration}), and comparisons of the use of PaLM \textit{versus} rule-based methods for hard negative (Sec.~\ref{sec:app:palm-comparison-neg}) and verb phrase (Sec.~\ref{sec:app:palm-comparison-vp}) generations.  

\vspace{+0.4cm}

\subsection{MSR-VTT}\label{sec:app:msr-vtt}
\begin{figure*}[!ht]
    \centering
    \includegraphics[width=0.90\textwidth]{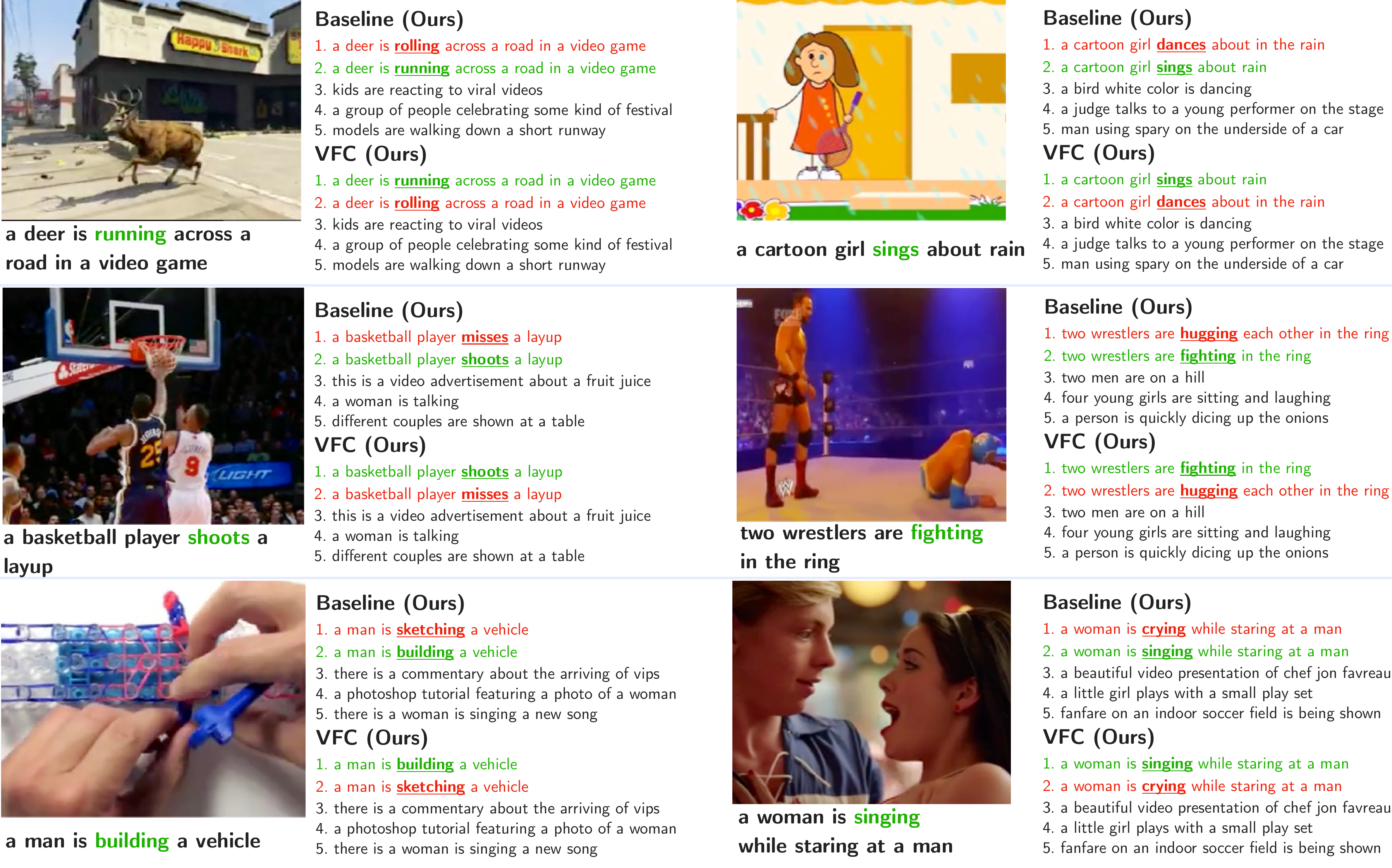}
     \includegraphics[width=0.90\textwidth]{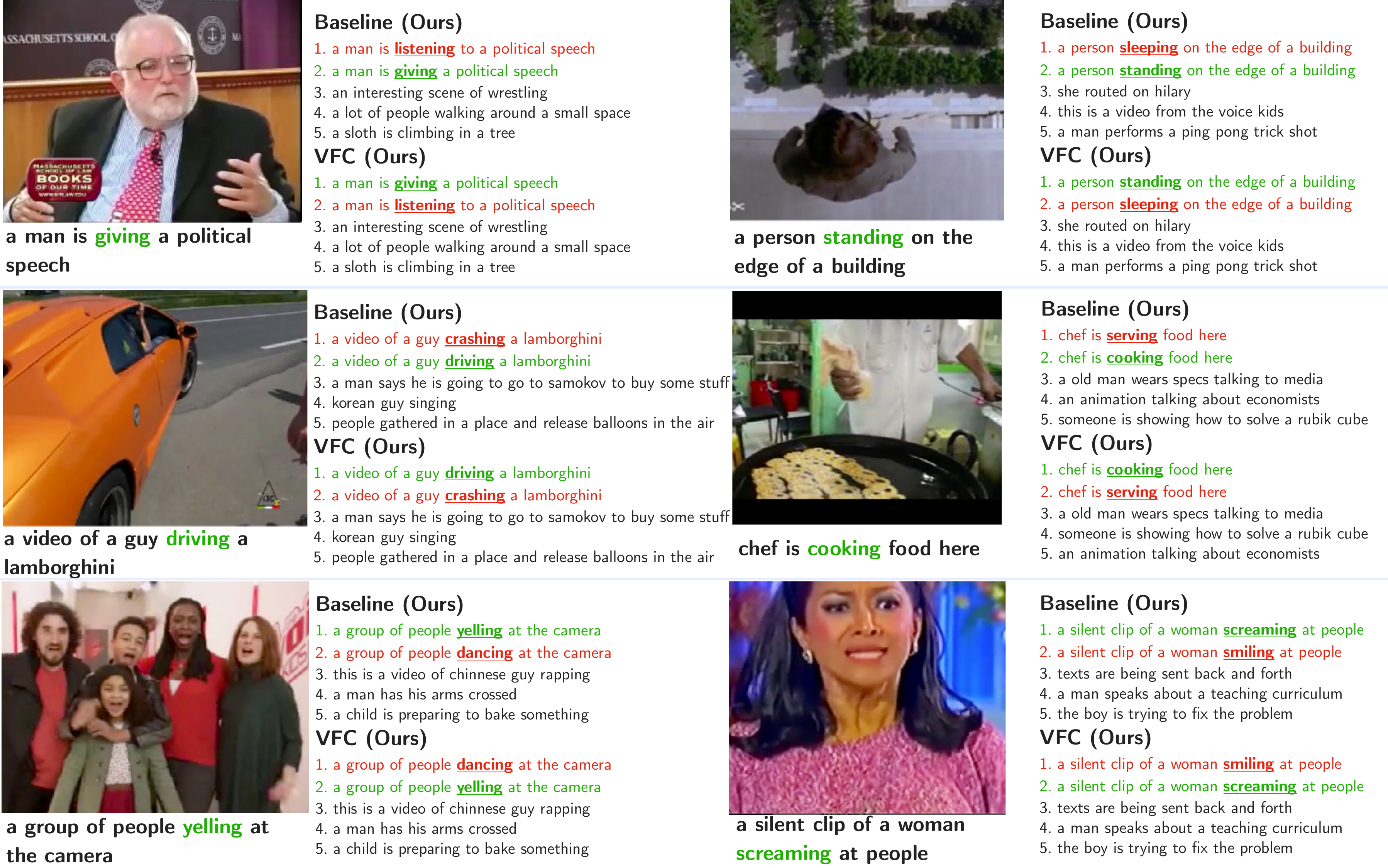}
    \caption{\textbf{MSR-VTT verb-focused benchmark:} We show qualitative examples from the Verb$_{H}$ \cite{park-etal-2022-exposing} multiple choice evaluation. For ease of visualisation, we only show a single frame per video. For each video sample, we show the 5 captions ranked in order of decreasing similarity for both our baseline and VFC models. We observe that the baseline model often mistakes the hard negative as matching the video; for example in the top left example, the caption `a deer is rolling across a road in a video game' is ranked higher than the correct answer `a deer is running across a road in a video game'. When training with hard negatives as in our VFC model, the model performance improves, retrieving the correct caption from the 5 options. On the bottom row, we show two failure cases where training with hard negatives causes the model to make a mistake; choosing the hard negative (`a group of people dancing at the camera', `a silent clip of a woman smiling at people') as the correct caption instead of (`a group of people yelling at the camera', `a silent clip of a woman screaming at people').}
    \label{fig:msrvtt}
\end{figure*}

We show qualitative examples from the Verb$_{H}$ \cite{park-etal-2022-exposing} multiple choice evaluation in Figure~\ref{fig:msrvtt}. For each video sample, we show the 5 captions ranked in order of decreasing similarity for both our baseline and VFC models. We observe that the baseline model often mistakes the hard negative as matching the video. This effect is reduced when training with hard negatives, as proposed in our VFC method, enabling the correct caption to be retrieved from the 5 options. In some rare cases, as shown on the last row, the baseline model is correct but training with hard negatives causes the hard negative to have highest similarity with the video. For example, the model incorrectly ranks `\textit{a silent clip of a woman \textbf{smiling} at people}' higher than `\textit{a silent clip of a woman \textbf{screaming} at people}'.

\clearpage

\clearpage

\begin{figure*}[!ht]
    \centering
    \includegraphics[width=0.9\textwidth]{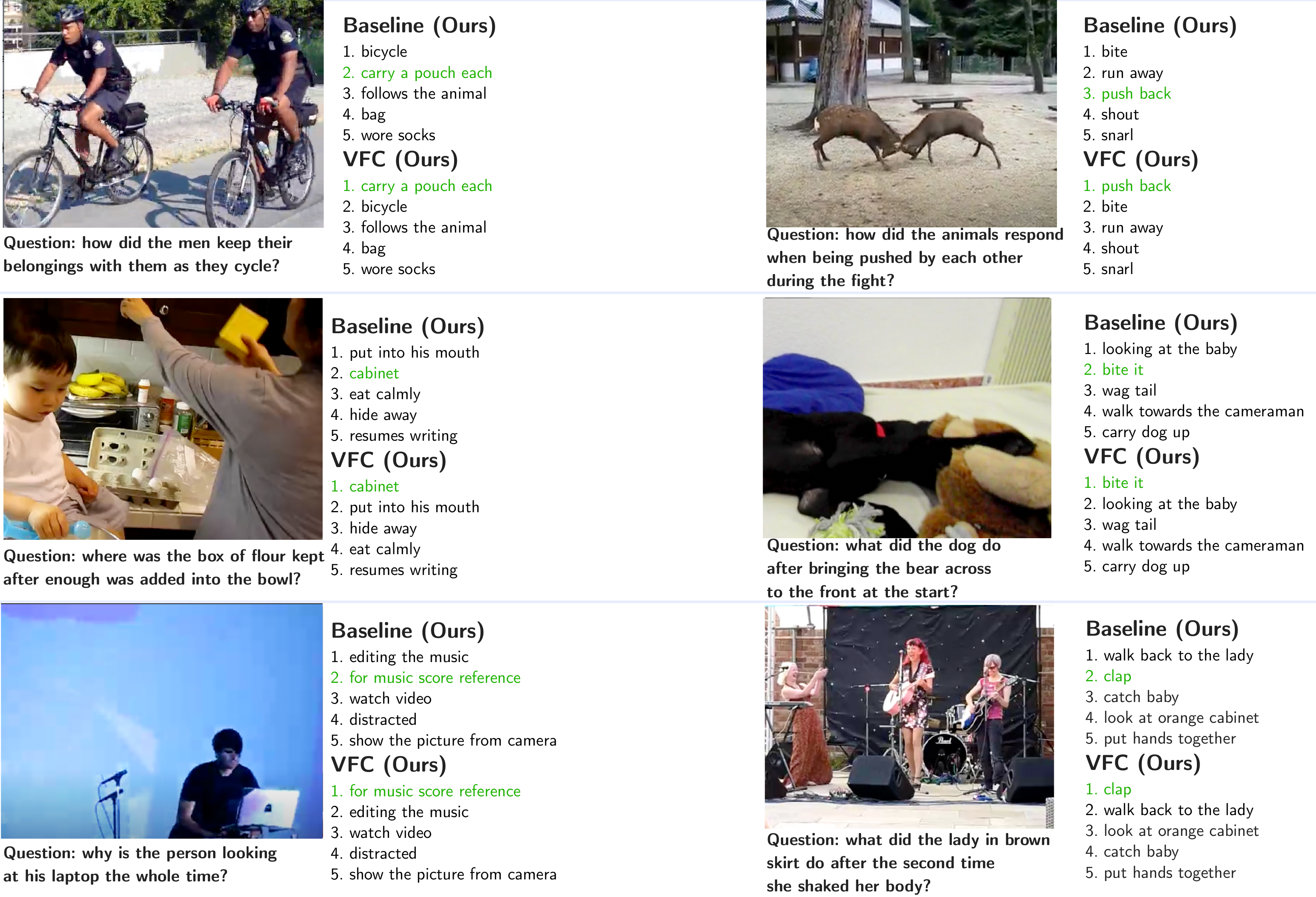}
    \vspace{0.5cm}
    \caption{\textbf{NEXT-QA:} We show qualitative examples from the ATP$_{hard}$ \cite{buch2022revisiting} multiple choice evaluation. For ease of visualisation, we only show a single frame per video. For each video sample, we show the 5 answers ranked in order of decreasing similarity for both our baseline and VFC models in a zero-shot setting. We observe that our VFC model improves performance, retrieving the correct answer from the 5 options more often.}
    \label{fig:nextqa}
\end{figure*}

\subsection{NEXT-QA}\label{sec:app:next-qa}
In Figure~\ref{fig:nextqa}, we show qualitative examples from the ATP$_{hard}$ \cite{buch2022revisiting} multiple choice evaluation. For each video sample, we show the 5 answers ranked in order of decreasing similarity for both our baseline and VFC models in a zero-shot setting. We observe that our VFC model improves performance, retrieving the correct answer from the 5 options more often.

\subsection{Kinetics-verb: Further analysis of calibration}\label{sec:app:calibration}
In Tab.~\ref{tab:calibration-supp}, we show further examples of confusion matrices comparing training performance with \textit{versus} without calibration on Kinetics-verb classes as in Tab.~\ref{tab:calibration} of the main paper. Once again, we observe that calibration reduces the effect of `attraction' points, which distort the feature space, by making $R_\omega$ the same for all verb phrase concepts.

\subsection{PaLM vs.~rule-based methods for hard negative generation}\label{sec:app:palm-comparison-neg}
\begin{figure}[t!]

    \centering
    \includegraphics[width=0.48\textwidth]{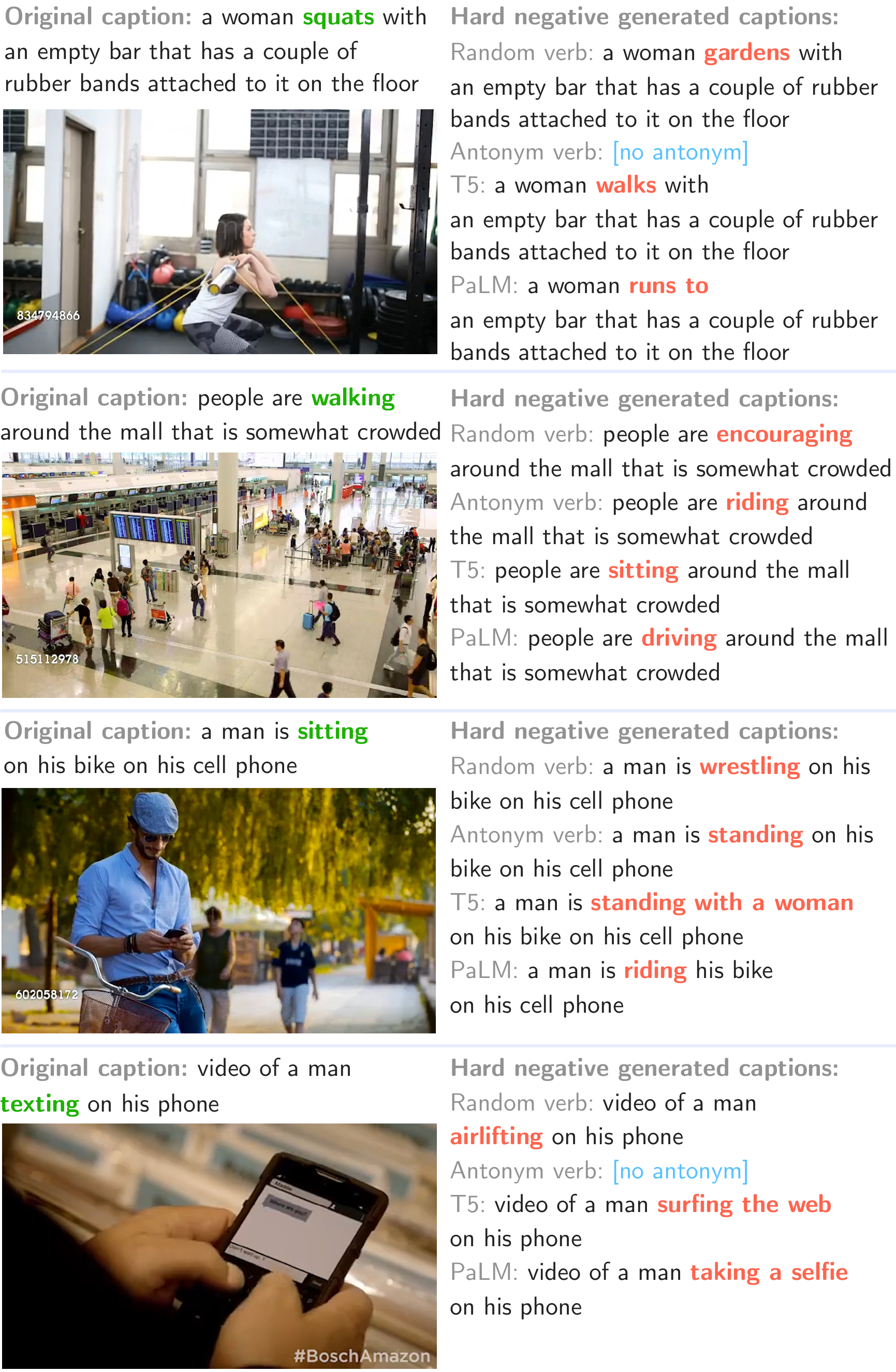}
    \caption{\textbf{PaLM vs.~rule-based methods for hard negative generation:} We compare hard verb negative caption generation using PaLM to T5 and rule-based methods such as replacing detected verbs by random verbs or antonym verbs.  We observe that randomly changing the verb often results in sentences which are linguistically and semantically incorrect, and that antonym verbs are often not present in NLTK~\cite{bird2009natural}. On the other hand, LLM based methods such as T5 and PaLM result in meaningful sentences. We note that LLM based methods can change more than just the verb: in the last row, replacing `texting' by `surfing the web' with T5 and `taking a selfie' with PaLM. In some cases, this can make it an easier negative: for example, in the third row, replacing `sitting' by `sitting with a woman' with T5.}
    \label{fig:palm-vs-rule-hn}
\end{figure}

In Fig.~\ref{fig:palm-vs-rule-hn}, we compare hard negative caption generation using PaLM to T5 and rule-based methods such as replacing detected verbs by random verbs or antonym verbs. We observe that LLM based methods result in linguistically and semantically viable sentences (which may not be guaranteed with random and antonym verb replacements). We also note that LLM based methods can change more than just the verb: (i) T5 and PaLM can replace the verb by a verb-noun pair and, (ii) PaLM can replace pronouns and determiners anywhere in the sentence (as opposed to T5 which can only replace the verb, see more details in Sec.~\ref{subsec:app:t5}), making the negative caption more linguistically correct.

\subsection{PaLM vs.~rule-based methods for verb phrase extraction}\label{sec:app:palm-comparison-vp}
\begin{figure}[H]
    \centering
    \includegraphics[width=0.48\textwidth]{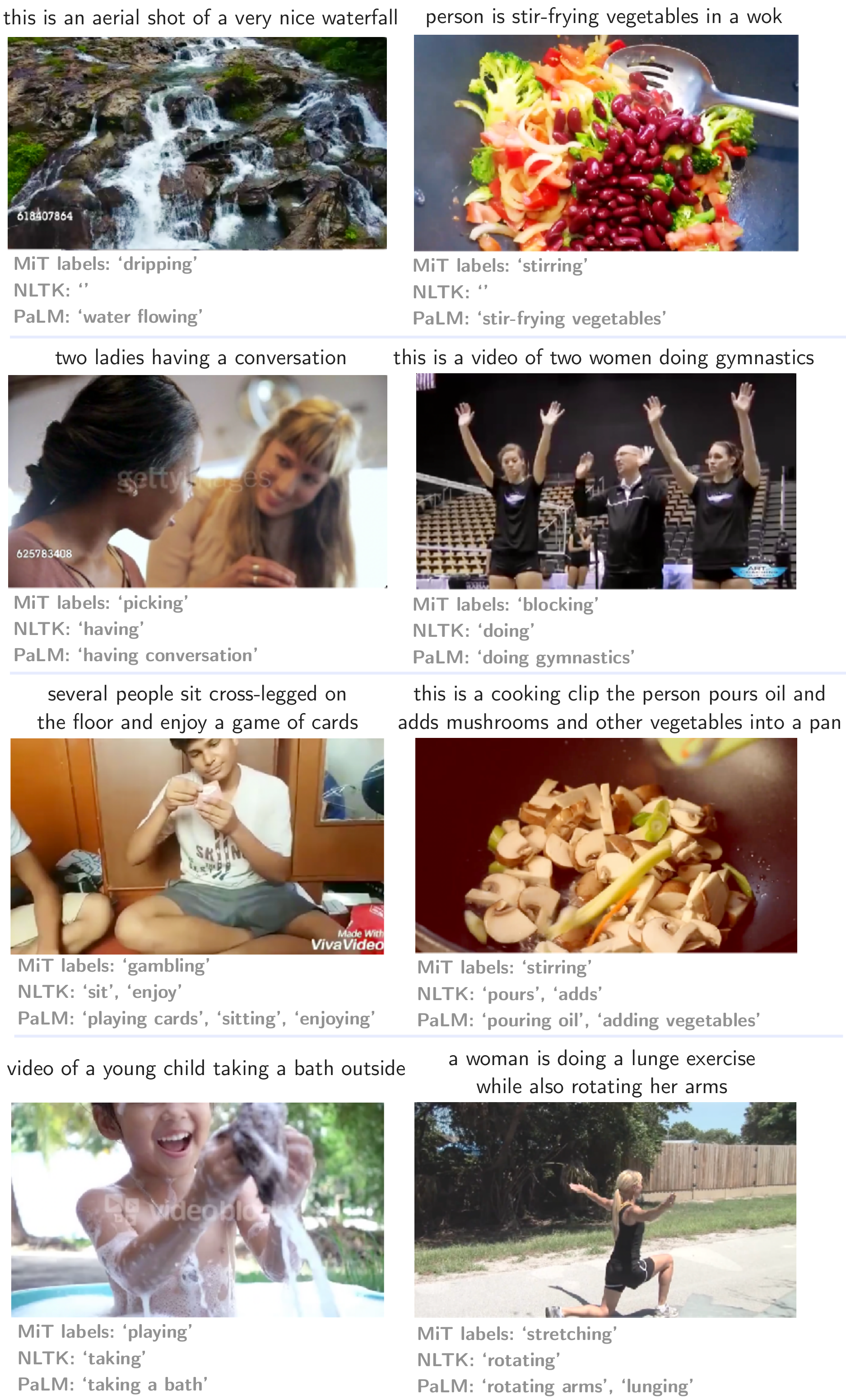}
    \caption{\textbf{PaLM vs.~rule-based methods for verb phrase extraction:} We compare verb phrase extraction using PaLM to:
(i)~using action labels for clips from the Moments in Time (MiT) dataset and (ii)~using a rule-based method such as NLTK~\cite{bird2009natural} to isolate verbs. In the top row, we show examples where NLTK outputs no label as a verb is not present in the sentence (first row, left) or is not detected (first row, right). In the second row, we show examples where extracting verbs with NLTK (e.g. `doing', `having') does not convey crucial information for understanding the action in the video. In the last two rows, we show examples where the MiT labels conceal valuable fine-grained action information in the video, whereas PaLM can recover this from the caption: (third row, left) the video is labelled as `gambling', PaLM extracts `playing cards'; (third row, right) the video is labelled as `stirring', PaLM extracts `pouring oil' and `adding vegetables'; (last row, left) the video is labelled as `playing', PaLM extracts `taking a bath'; (last row, right) the video is labelled as `streching', PaLM extracts `rotating arms' and `lunging'. Overall, our PaLM method of extracting verbs from captions performs best qualitatively and quantitatively (as shown in Tab.~\ref{fig:multi_ablat} (right) of the main paper).}
    \label{fig:palm-vs-rule-vp}
\end{figure}

In Fig.~\ref{fig:palm-vs-rule-vp}, we compare verb phrase extraction using PaLM to:
(i) using action labels for clips from the Moments in Time (MiT) dataset (these are available as SMiT data inherits from MiT~\cite{monfortmoments}) and
(ii) using a rule-based method (NLTK~\cite{bird2009natural}) to isolate verbs. 
We observe that using PaLM outperforms both: (i) MiT action labels can be general and conceal fine-grained action information in the video which can improve verb understanding, (ii) NLTK has difficulties extracting all verbs in a sentence, and can often mistake them for nouns; NLTK also cannot extract a verb phrase when a verb is not present in the sentence (e.g. for the caption `this is an aerial shot of a very nice waterfall', NLTK extracts no verb phrase while PaLM extracts `water flowing'); finally, our NLTK approach does not extract verb-noun pairs (e.g. for the caption `this is a video of two women who are doing gymnastics', NLTK extracts `doing' while PaLM extracts `doing gymnastics') -- this can be crucial for understanding the action in the video. We note that although NLTK could be used to extract verbs and nouns independently through PoS tagging, correctly assigning nouns to the matching verb is not always robust for long, complex sentences as in SMiT. Indeed, the average length of a sentence in SMiT is 18 words. 

\begin{table}[ht]
\vspace{0.5cm}
\setlength{\tabcolsep}{0.1pt}
\centering
\begin{tabular}{l r r}
\toprule
& ~~~~~~~~~~~~\small{w/o calibration} & ~~~~~~~~~~~~~\small{w/ calibration} \\
\vspace{+0.1cm}
\small{$R_{\omega}~~~\propto$} & \multirow{2}{*}{\includegraphics[width=0.25\linewidth]{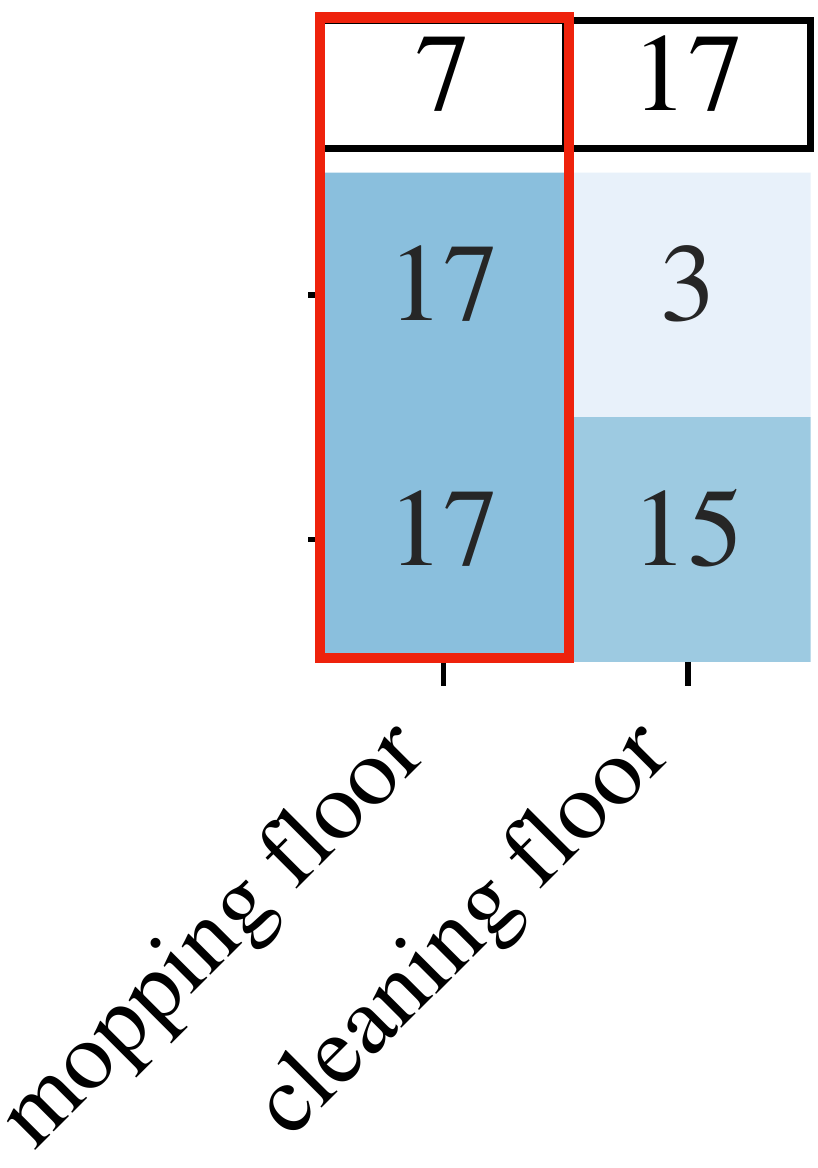}} & \multirow{5}{*}{\includegraphics[width=0.25\linewidth]{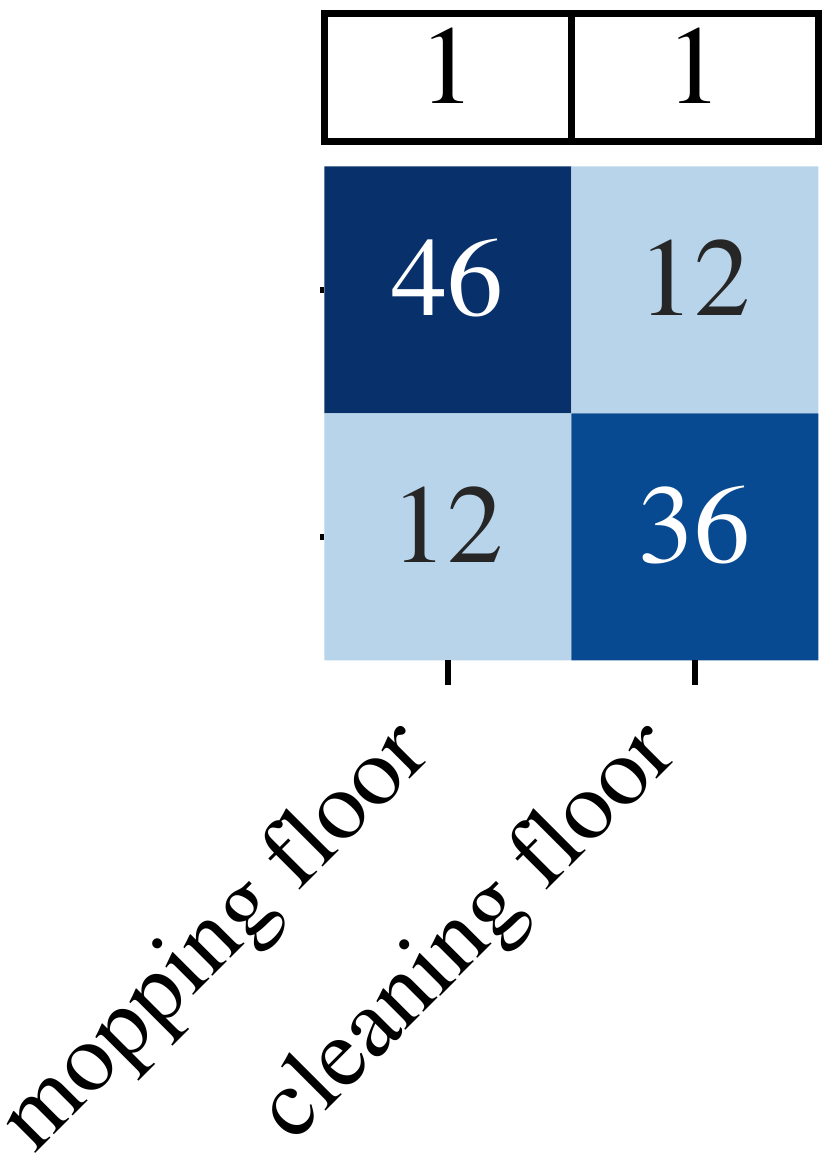}}\\
\vspace{+0.1cm}
\small{mopping floor} & \\
\vspace{+0.1cm}
\small{cleaning floor} & \\
\\
\\
\\
\\
\midrule
\small{$R_{\omega}~~~\propto$} & \multirow{2}{*}{\includegraphics[width=0.27\linewidth]{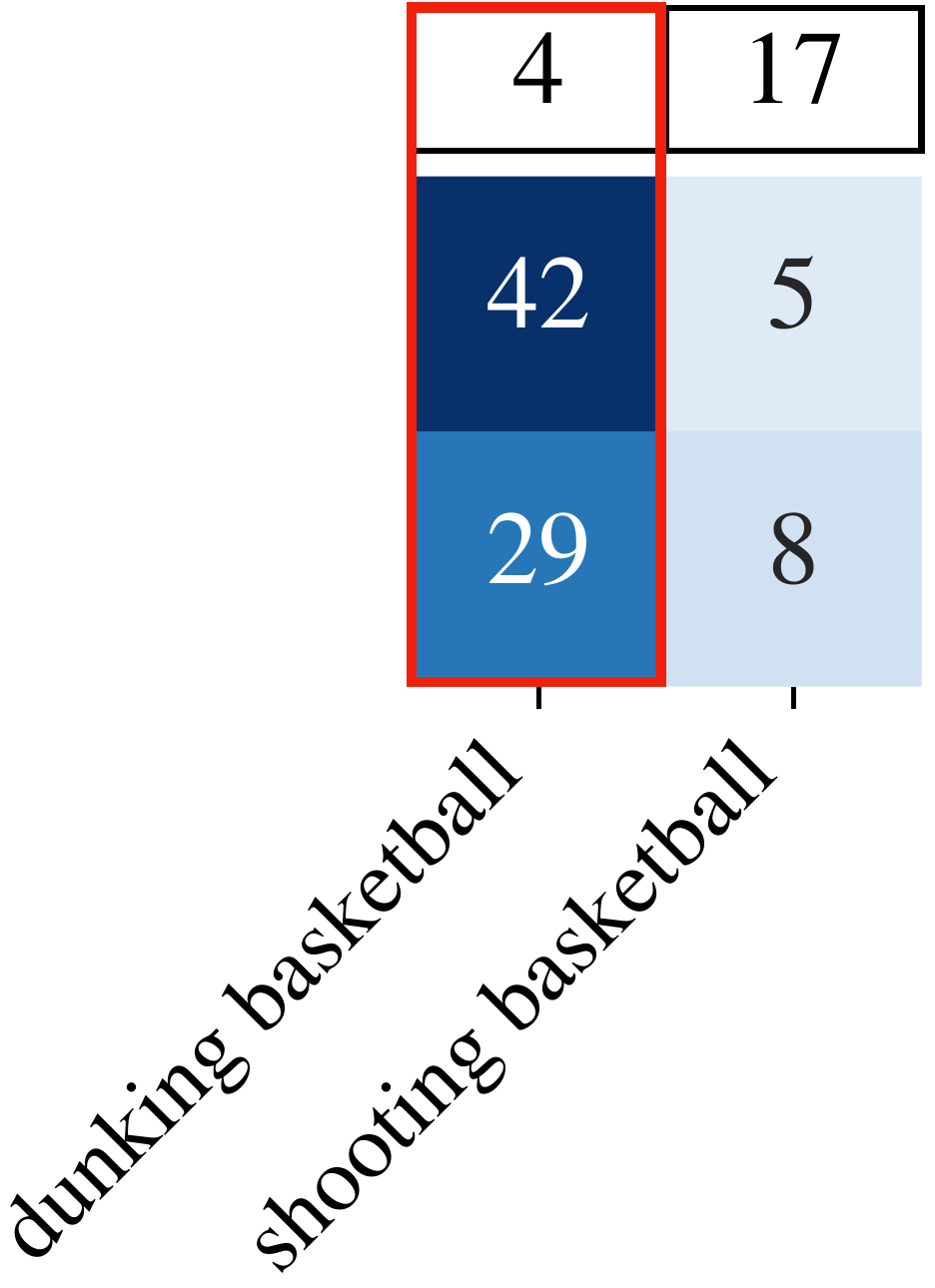}} & \multirow{5}{*}{\includegraphics[width=0.27\linewidth]{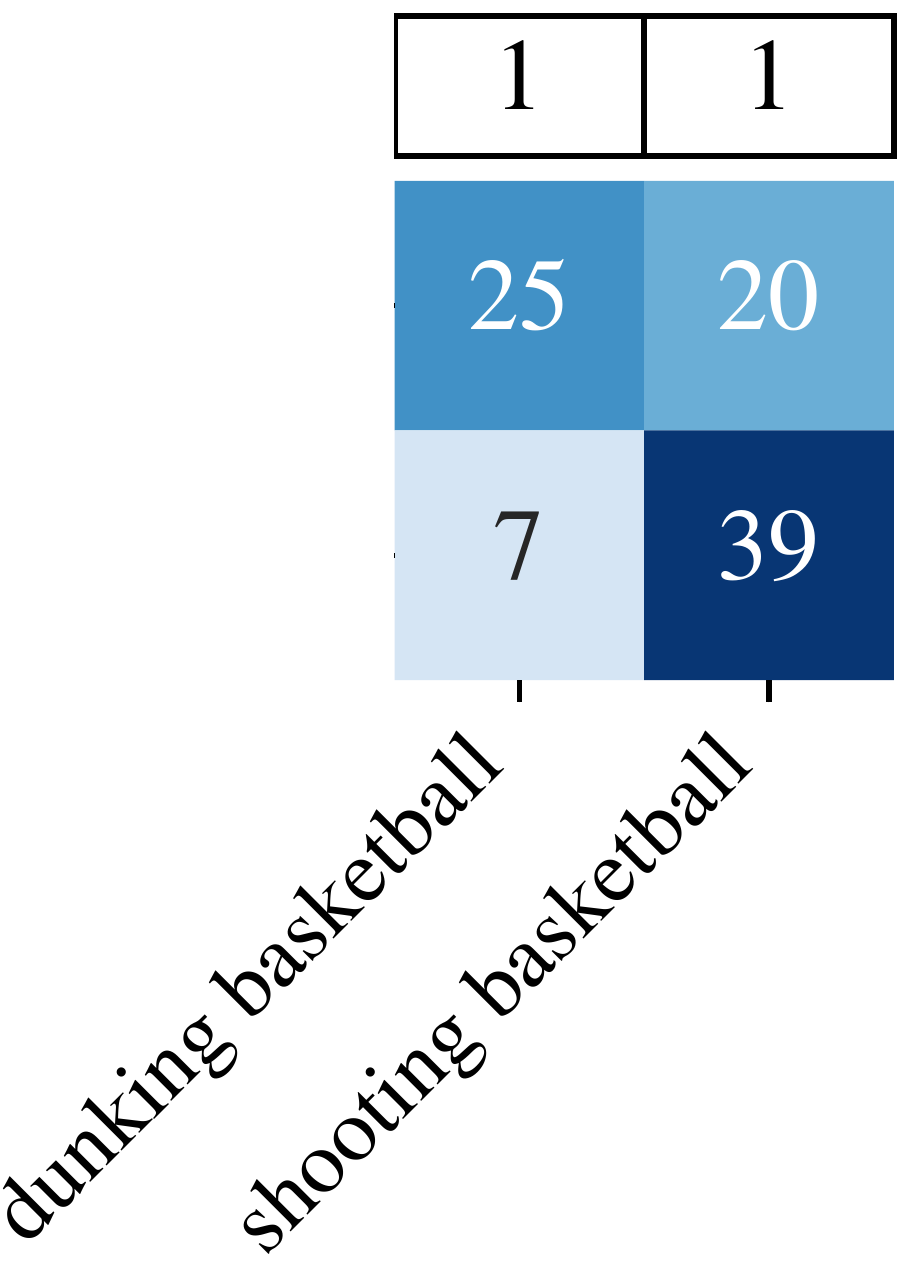}}\\
\vspace{+0.1cm}
\small{dunking basketball} & \\
\vspace{+0.1cm}
\small{shooting basketball} & \\
\\
\\
\\
\\
\midrule
\small{$R_{\omega}~~~\propto$} & \multirow{2}{*}{\includegraphics[width=0.21\linewidth]{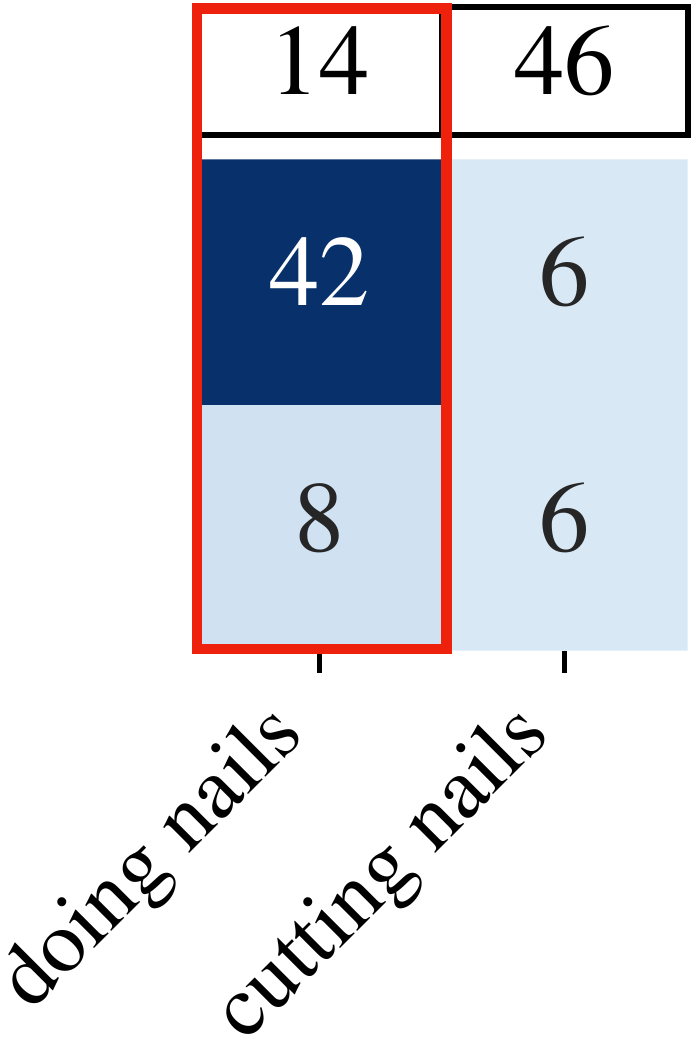}} & \multirow{5}{*}{\includegraphics[width=0.21\linewidth]{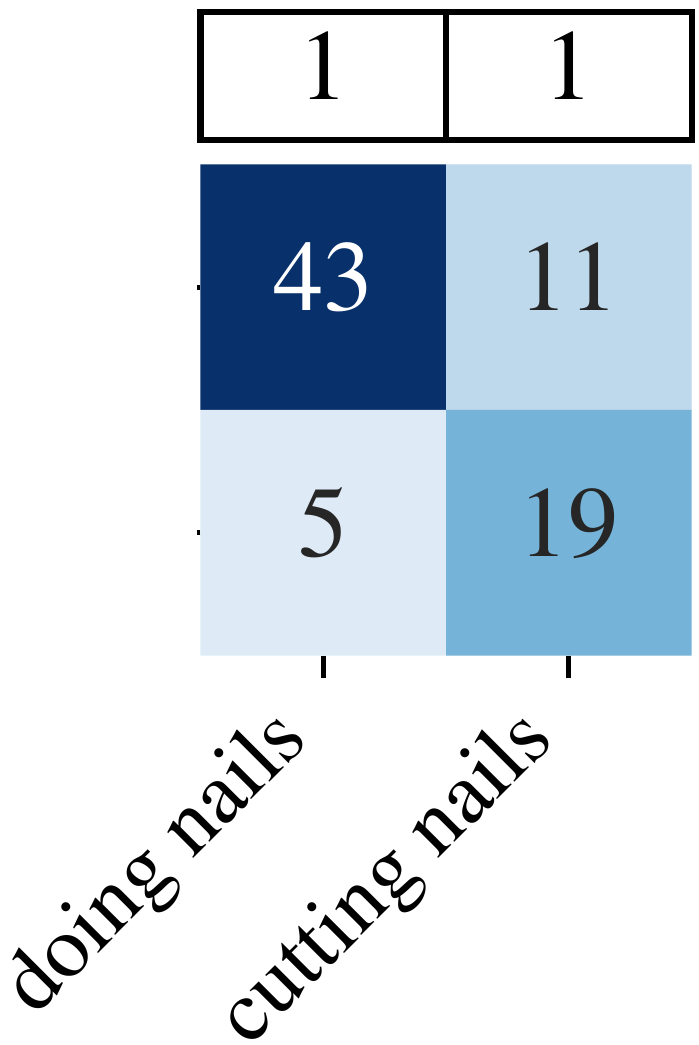}}\\
\vspace{+0.1cm}
\small{doing nails} & \\
\vspace{+0.1cm}
\small{cutting nails} & \\
\vspace{0.7cm}
\end{tabular}
    \caption{
    \textbf{Confusion matrix for Kinetics-verb classes.}
Without proper calibration, the verb phrases `mopping floor', `dunking basketball', `doing nails' become highly attractive in the video-text feature space. Our calibration mechanism alleviates this issue by making the ratio $R_\omega$ independent of verb phrases (see details in Sec.~\ref{sec:vfc} of the main paper).
    }
    \label{tab:calibration-supp}
\end{table}

\section{Baselines \& Implementation details}\label{sec:app:details}

In this section, we present detailed descriptions of baselines (Sec.~\ref{subsec:app:baseline}), the CLIP4CLIP~\cite{Luo2021CLIP4Clip} architecture used in all our experiments (Sec.~\ref{subsec:app:clip4clip}), fine-tuning (Sec.~\ref{subsec:app:finetuning}) and evaluation protocols (Sec.~\ref{subsec:app:eval_proto}), the PaLM prompting procedure (Sec.~\ref{subsec:app:palm}), the T5 hard negative generation process (Sec.~\ref{subsec:app:t5}), and the Kinetics-verb split we propose (Sec.~\ref{subsec:app:kinetics-verb}). 

\subsection{Baselines}\label{subsec:app:baseline}

We describe in more detail baselines presented in the main paper for MSR-VTT, NEXT-QA, Kinetics-400 and SVO-Probes.

\clearpage

\begin{figure*}[ht!]
    \centering
    \includegraphics[width=\textwidth]{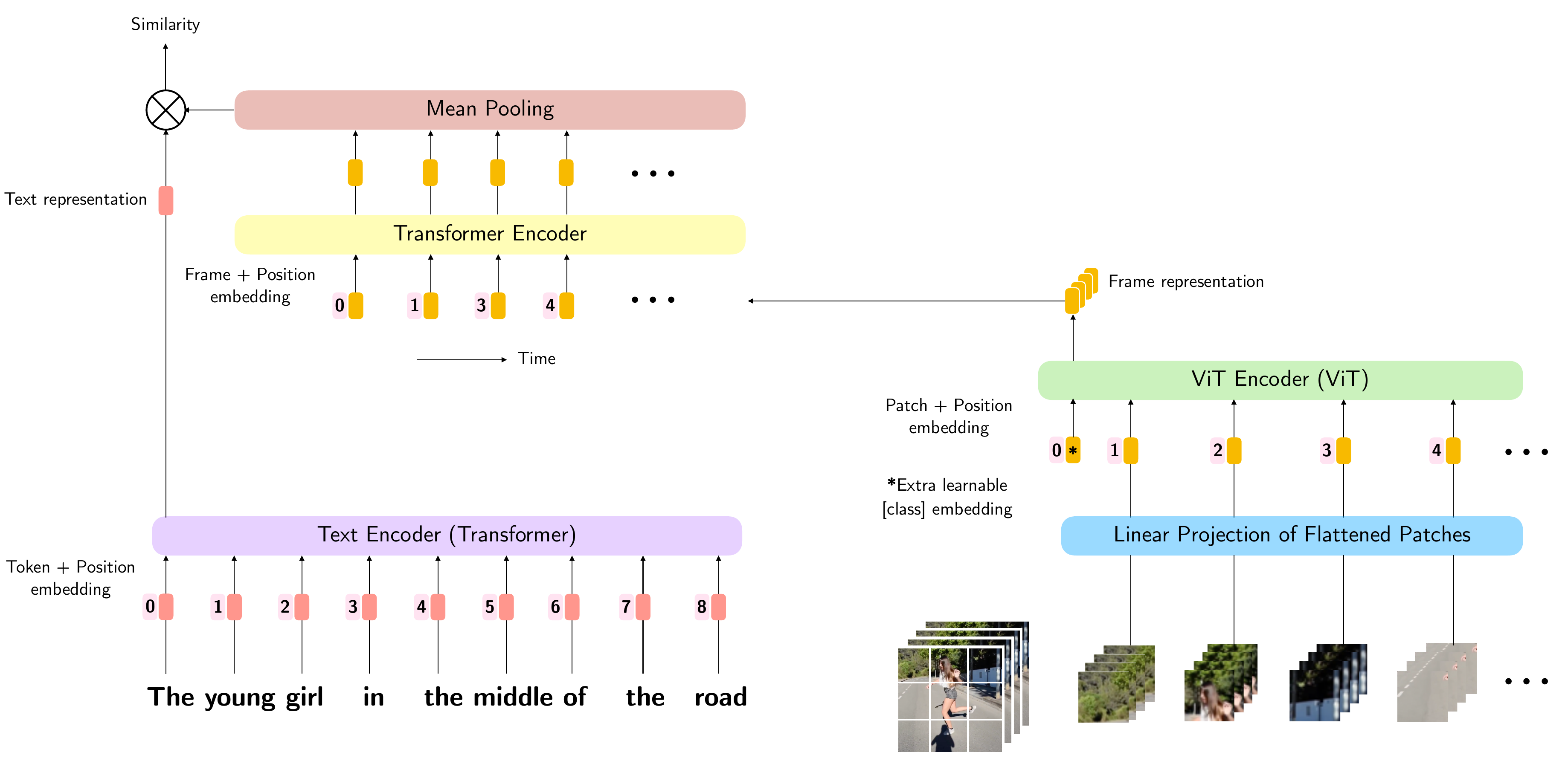}
    \caption{\textbf{CLIP4CLIP Architecture:} Figure adapted from~\cite{Luo2021CLIP4Clip}. The model consists of a video encoder, text encoder and similarity calculator. Each frame is passed through ViT to obtain a frame representation at the output of the [class] token. The $T$ frame representations are then passed through a Transformer for sequence modelling and averaged with a mean pooling operation to obtain a video-level representation. The video representation is then compared to the text representation through the cosine similarity. }
    \label{fig:qualitative}
\end{figure*}

\noindent{\textbf{MSR-VTT.}} We show the performance of VideoCLIP~\cite{xu-etal-2021-videoclip}, CLIP~\cite{Radford2021CLIP} and InternVideo~\cite{wang2022internvideo} zero-shot. VideoCLIP trains a transformer for video and text by contrasting temporally overlapping positive video-text pairs with hard negatives from nearest neighbor retrieval. More details for the CLIP baseline can be found in~\cite{Radford2021CLIP}. InternVideo explores jointly using masked video modelling and video-language contrastive learning as pretraining objectives. In the fine-tuned setting, we compare to ClipBERT~\cite{lei2021less}, MMT~\cite{gabeur2020mmt}, VideoCLIP~\cite{xu-etal-2021-videoclip}, C4CL-mP~\cite{park-etal-2022-exposing}. ClipBERT focuses on sparse training to reduce video processing overhead and applying image-text pretraining for video-text tasks. MMT uses a multi-modal transformer to encode video and BERT~\cite{devlin-etal-2019-bert} for text. C4CL-mP corresponds to the CLIP4CLIP~\cite{Luo2021CLIP4Clip} reimplementation by Park et al.~\cite{park-etal-2022-exposing} with just mean pooling (without any Transformer Encoder for temporal modelling of frames).

\noindent{\textbf{NEXT-QA.}}  We show the performance of CLIP zero-shot. More details for this baseline can be found in~\cite{Radford2021CLIP}. In the fine-tuned setting, we compare to HGA~\cite{jiang2020reasoning}: a deep heterogenous graph network which aligns inter- and intra- modality information (appearance, motion and text) to reason and answer the question. We also compare to ATP, Temp[ATP] and Temp[ATP]+ATP from~\cite{buch2022revisiting}. ATP consists of a Transformer which learns to select a single (frozen) CLIP frame embedding from a video, given the sequence of video frame embeddings and question embedding, for the task of video question answering. For training, they use a cross entropy loss over the answer set. Temp[ATP] is an extension of ATP, where the video is first partitioned into $k$ clips, and a single frame embedding is selected using ATP from each clip. These $k$ frame embeddings are then aggregated to a video-level representation using a Transformer, before being passed to the downstream task. Temp[ATP]+ATP corresponds to an ensemble of both ATP and Temp[ATP]. Finally we compare to VGT~\cite{xiao2022video}, which consists of a video graph transformer that explicitly encodes objects, relations and dynamics. VGT also uses disentangled video and text Transformers to better measure relevance between video and text.

\noindent{\textbf{Kinetics-400.}} We show the performance of Flamingo~\cite{flamingo}, ActionCLIP~\cite{actionclip} and CLIP~\cite{Radford2021CLIP} zero-shot. Flamingo is a visual-language model, which leverages pretrained vision and language models and bridges them effectively by using gated cross-attention and dense layers.  ActionCLIP~\cite{actionclip} reformulates action recognition into a video-text matching problem within a multimodal contrastive learning framework. More details for CLIP can be found in~\cite{Radford2021CLIP}.

\noindent{\textbf{SVO Probes.}} We show the performance of No-MRM–MMT (the best performing model in \cite{hendricks2021probing}) and CLIP~\cite{Radford2021CLIP} zero-shot. No-MRM-MMT corresponds to a multi-modal transformer (similar to the ViLBERT~\cite{vilbert} architecture) with a masked language modeling loss (MLM), an image-text matching loss (ITM) that classifies if an image-sentence pair are matching, but no masked region modeling loss (MRM). More details for the CLIP baseline can be found in~\cite{Radford2021CLIP}.

\subsection{CLIP4CLIP Architecture}\label{subsec:app:clip4clip}

Here, we describe the CLIP4CLIP network architecture~\cite{Luo2021CLIP4Clip}, illustrated in Figure~\ref{fig:qualitative}, used in all our experiments. This architecture consists of three components: a video encoder, a text encoder, and a similarity calculator. We describe each component in detail next. We note that all three components are fine-tuned in all our experiments.

\noindent \textbf{Video encoder}. The pretrained CLIP (ViT-B/32)~\cite{Radford2021CLIP} image encoder is used to obtain frame representations. Specifically, video frames are first sampled from the video and reshaped into a sequence of flattened 2D patches. These patches are then linearly projected to 1D tokens before being inputted to ViT~\cite{dosovitskiy2020vit}, a 12-layer Transformer. The output from the [class] token is used as the video frame representation: given $T$ input frames, we obtain $T$ frame representations. In practice, we select 32 frames (with initial resolution $256 \times 256$, of which augmented crops of size $224 \times 224$ are taken) in a video at 25fps, at a stride of 2 frames for training. 

\noindent \textbf{Text encoder}. The CLIP pretrained text encoder is used to embed the caption. It corresponds to a 12-layer Transformer model; further details can be found in~\cite{Radford2021CLIP}. 

\noindent \textbf{Similarity calculator}. The goal is to learn a function to calculate the similarity between video-text pairs inputted to the model in such a way that video-text pairs which \textit{match} have a high similarity, and otherwise have a low similarity.  Therefore, we ultimately want to compare a text and video-clip representation. The ViT encoder outputs a representation for each of the sequence of frames  without any temporal modelling. We therefore first pass these frame embeddings (along with temporal positional embeddings) through a 4-layer Transformer encoder. We then apply a mean-pooling operation to the new frame embeddings to obtain a video-level representation. Finally, we calculate the cosine similarity between the video and text representations.

Following the protocol in~\cite{Luo2021CLIP4Clip}, the positional embeddings in the similarity calculator are initialised by repeating the position embedding from CLIP's text encoder. The Transformer encoder is initialised by the corresponding layers' weight of the pretrained CLIP image encoder. The rest is randomly initialised.

\subsection{Fine-tuning details}\label{subsec:app:finetuning}

\noindent \textbf{MSR-VTT}. When fine-tuning on MSR-VTT, we use the 9K and 7K training split for the retrieval and multi-choice settings respectively. For the 9K split, we train for 100 epochs with a base learning rate of 1e-7, a weight decay of 1e-2 and temperature of 5e-3. For the 7K split, we train for 100 epochs with a base learning rate of 1e-7, a weight decay of 1e-2 and temperature of 5e-3. For both settings, we train with the hard negative contrastive loss and discard the verb phrase loss. Indeed, we use PaLM to generate hard negative captions for MSR-VTT, since it is a video-text retrieval dataset, similarly to SMiT. We sample 32 frames per video at 25~fps with a stride of 14.

\noindent \textbf{NEXT-QA}. For fine-tuning on NEXT-QA, we concatenate the question and answer pairs before passing them through the CLIP4CLIP text tower. We continue using the hard-negative cross-modal contrastive loss during fine-tuning, treating the four incorrect question-answer pairs as hard negatives. We discard the verb phrase loss. We train for 100 epochs with a base learning rate of 1e-6, a weight decay of 5e-2 and temperature of 1e-3. We maintain a batch size of 256. We sample 32 frames per video at 25~fps with a stride of 24.

\subsection{Evaluation protocols}\label{subsec:app:eval_proto}

\noindent \textbf{MSR-VTT}. For the standard setting, we evaluate text-to-video retrieval (R@1) on the 1K validation split and 3K Random MC. In the former, the model must associate the text to the correct video, among 1000 videos. For the latter, the model must associate the video to the right caption, among 5 captions, where the 4 negative captions are randomly chosen from other videos. For our verb-focused setting, we use the Verb$_{H}$ multiple choice (MC) validation split from \cite{park-etal-2022-exposing}. Verb$_{H}$ MC covers a subset of the videos in the 3K Random MC split, with 2,554 video-text instances, but the task is harder. In the Verb$_{H}$ MC setting, one of random negative captions is replaced by a \textit{hard verb negative}, where the correct sentence's verb has been modified manually in such a way that the new sentence is inconsistent with the video. We mark the model prediction as correct if the ground truth sentence among the 5 captions has the highest similarity score with the video. We sample 32 frames per video at 25 fps with a stride of 14.

\noindent \textbf{Kinetics-400}. We follow~\cite{Radford2021CLIP} to evaluate classification in a zero-shot setting: we feed in all class labels (without any prompt) to the text tower and mark the prediction as correct if the correct label has the highest similarity with the video. For the `Kinetics-verb' split, we restrict the evaluation to 97 classes which we manually identify as requiring verb understanding (see Sec.~\ref{subsec:app:kinetics-verb}). We note that we still feed in all 400 class labels for measuring classification on Kinetics-verb. We sample 32 frames per video at 25 fps with a stride of 14.

\noindent \textbf{NEXT-QA}. We concatenate the question and answer pairs before passing them through the CLIP4CLIP text tower. We mark the model prediction as correct if the correct question-answer pair among the 5 options has the highest similarity score with the video. We sample 32 frames per video at 25~fps with a stride of 24.

\noindent \textbf{SVO probes}. This is an image-text benchmark~\cite{hendricks2021probing}, specifically designed to measure progress in verb understanding. We evaluate our baseline and VFC framework on a subset of 12,936 images from the original 14,102 images since some images are no longer accessible (the corresponding urls are corrupted). In~\cite{hendricks2021probing}, the authors calculate the accuracy of positive and negative image-text pairs: they pass image-text pairs through their model and label an image–sentence pair as negative if the classifier output is $< 0.5$ and
positive otherwise. Our model confidences are calibrated differently, therefore we instead report Average Precision (AP). To evaluate on this dataset, we simply replicate the image 32 times as input to our video model.

\subsection{PaLM prompting}\label{subsec:app:palm}

\noindent\textbf{PaLM hard negative generation.} We include below our full prompt template for automatic generation of hard negatives. We insert the caption for which we want to generate hard verb negatives at \textbf{\{input caption\}}. \\
\noindent\rule{8.5cm}{0.4pt}
\textit{In this task, you are given an input sentence. Your job is to tell me 10 output sentences with a different meaning by only changing the action verbs. \\
Input: A man walks up to a woman holding an umbrella in a garden. \\
Outputs: \\
1) A man jumps up to a woman throwing an umbrella in a garden. \\
2) A man runs up to a woman opening an umbrella in a garden. \\
3) A man walks away from a woman buying an umbrella in a garden. \\
4) A man throws up on a woman carrying an umbrella in a garden. \\
5) A man punches a woman swinging an umbrella in a garden. \\
6) A man sits with a woman wrapping up her umbrella in a garden. \\
7) A man talks to a woman closing an umbrella in a garden. \\
8) A man flirts with a woman playing with an umbrella in a garden. \\
9) A man skips to a woman leaning on her umbrella in a garden. \\
10) A man sprints to a man losing her umbrella in a garden. \\
Input: Surfers ride the waves in an ocean. 
Outputs: \\ 
1) Surfers get hit by the waves in an ocean. \\
2) Surfers swimming in the waves in an ocean. \\
3) Surfers meditating by the waves in an ocean. \\
4) Surfers drowning in the waves in an ocean. \\
5) Surfers asking for help in the waves in an ocean. \\
6) Surfers teaming up in the waves in an ocean. \\
7) Surfers snorkeling in the waves in the ocean. \\
8) Surfers taking photos by the waves in the ocean. \\
9) Surfers getting ready to go into the waves in the ocean. \\
10) Surfers st}\textit{retching by the waves in the ocean. \\
Input: A dentist holds the replica of a human mouth he shows how important flossing your teeth is. \\
Outputs:\\
1) A dentist cleans the replica of a human mouth he presents how unimportant flossing your teeth is. \\
2) A dentist breaks the replica of a human mouth he screams how important flossing your teeth is.\\
3) A dentist fixes the replica of a human mouth he says how important flossing your teeth is. \\
4) A dentist buys the replica of a human mouth he explains how important brushing your teeth is. \\
5) A dentist plays with the replica of a human mouth he remembers about how important washing your teeth is. \\
6) A dentist tidies the replica of a human mouth he rambles on about how important breaking your teeth is. \\
7) A dentist rotates the replica of a human mouth he presents how important fracturing your teeth is. \\
8) A dentist places on his legs the replica of a human mouth he shows how important flossing your teeth is.\\
9) A dentist searches for the replica of a human mouth he shows how important grinding your teeth is. \\
10) A dentist picks up the replica of a human mouth he presents how important whitening your teeth is. \\
Input: Looks like a band playing on the stage and perhaps Community Center and people gathered around watching. \\
Outputs: \\
1) Looks like a band fighting on the stage and perhaps Community Center and people gathered around crying. \\
2) Looks like a band dancing on the stage and perhaps Community Center and people gathered around smiling. \\
3) Looks like a band singing on the stage and perhaps Community Center and people gathered around filming. \\
4) Looks like a band bowing on the stage and perhaps Community Center and people gathered around clapping.\\
5) Looks like a band making a speech on the stage and perhaps Community Center and people gathered around listening. \\
6) Looks like a band laughing on the stage and perhaps Community Center and people gathered around cheering.\\
7) Looks like a band working on the stage and perhaps Community Center and people gathered around standing.\\
8) Looks like a band holding hands on the stage and perhaps Community Center and people gathered around praying. \\
9) Looks like a band jumping on the stage and perhaps Community Center and people gathered around encouraging.\\
10) Looks like a band yelling on the stage and perhaps Community Center and people gathered around watching.\\
Input: \textbf{\{input caption\}}\\
Outputs: \\}
\noindent\rule{8.5cm}{0.4pt}

\noindent\textbf{PaLM verb phrase extraction.} We use PaLM to extract verb phrases from the original caption, where a verb phrase can correspond to a single verb or a verb-noun pair depending on the caption. We use PaLM-540B with output sequence length 256, beam size of 4,  and temperature of 0.2. We post-process the outputs by removing text after any newline character. We include our full prompt template for automatic extraction of verb phrases below. We insert the caption for which we want to extract a verb phrase at \textbf{\{input caption\}}. \\
\noindent\rule{8.5cm}{0.4pt}
\textit{In this task, you are given an input sentence. Your job is to output the action verb phrases. \\
Input: the young girl in the middle of the road she is dancing. \\
Output:  [`dancing'] \\
Input: a city area can be seen that has people in the walkways of runways. \\
Output: [] \\
Input: this is a video of a birthday and she has a green colored dress and they are cutting a cake there's a clown on the side and the parents seem to be clap. \\
Output: [`cutting cake', `clapping'] \\
Input: one woman is talking to the camera about being safe he has a shirt with pal pal on it in the greenery behind her.  \\
Output: [`talking to camera'] \\
Input: a bicycle with a specialized back wheel slides along a wet paper. \\
Output: [`sliding'] \\
Input: a person clicking an object that is connected to a speaker. \\
Output: [`clicking'] \\
Input: it's a video of a football game and one of the blue team is throwing the football really far into the endzone. \\
Output: [`throwing football'] \\
Input: this is a video of someone filing their nails. \\
Output: [`filing nails'] \\
Input: airplane with the words British Airways can be seen over top. \\
Output: [] \\
Input: man sitting standing at the front of the room is giving speech and asking an audience if they've ever heard of a specific song. \\
Output: [`standing', `giving speech', `asking'] \\
Input: it shows a video of a man talking on the phone yeah glasses and has a black phone. \\
Output: [`talking on phone'] \\
Input: hitchhiker is on the side of the road by a truck stop pulling a sign that says North. \\
Output: [`pulling a sign'] \\
Input: this is a video of a man on a ladder the man is cutting down a tree branch the man is wearing red. \\
Output: [`cutting tree'] \\
Input: on an indoor gym on a hard Brown meth there's a man young man with a barbell with lots of heavy weights on each side and he has it over his head stiff arm straight arm going to be and then he drops it on the floor while he does so you can hear the clanking of the weight that they smack against each other. \\
Output: [`dropping'] \\
Input: he is using a large chainsaw to cut inside of a tree branch. \\
Output: [`cutting tree'] \\
Input: I meant stacking up his cups for cup stacking concentration for a party. \\
Output: [`stacking cups'] \\
Input: a large field shown with garbage and water flowing through it. \\
Output: [`water flowing'] \\
Input: a washing machine washes the clothes. \\
Output: [`washing clothes'] \\
Input: \textbf{\{input caption\}}\\
Output: \\}
\noindent\rule{8.5cm}{0.4pt}

\noindent\textbf{PaLM positive generation.} We use PaLM to generate positive sentences where the verb in the original caption is changed to a synonym verb, but the remaining context is unchanged. We use PaLM-540B with output sequence length 512, beam size of 1, and temperature of 0.7. We post-process the outputs by removing text after any newline character and by filtering out candidates which contain the same verbs as the original caption. We include our full prompt template for automatic generation of positives below. We insert the caption for which we want to generate a positive sentence at \textbf{\{input caption\}}. \\
\noindent\rule{8.5cm}{0.4pt}
\textit{In this task, you are given an input sentence. Your job is to tell me 10 output sentences with the same meaning by only changing the action verbs. \\
Input: A man walks up to a woman holding an umbrella in a garden. \\
Outputs: \\
 1) A man strolls up to a woman holding an umbrella in a garden.  \\
 2) A man marches up to a woman holding an umbrella in a garden.  \\
 3) A man strides up to a woman holding an umbrella in a garden. \\ 
 4) A man wanders up to on a woman carrying an umbrella in a garden. \\
 5) A man tramps up to a woman holding an umbrella in a garden. \\
 6) A man steps up to with a woman holding an umbrella in a garden. \\
 7) A man wanders up to a woman holding an umbrella in a garden. \\ 
 8) A man treads up to a woman holding an umbrella in a garden. \\
 9) A man truges up to a woman holding an umbrella in a garden. \\
 10) A man treaks to a woman holding her umbrella in a garden. \\
Input: A dentist holds the replica of a human mouth he shows how important flossing your teeth is. \\
Outputs: \\ 
1) A dentist grasps the replica of a human mouth he shows how important flossing your teeth is. \\
2) A dentist carries the replica of a human mouth he shows how important flossing your teeth is. \\
3) A dentist clutches the replica of a human mouth he shows how important flossing your teeth is. \\
4) A dentist grips the replica of a human mouth he shows how important flossing your teeth is. \\
5) A dentist holds the replica of a human mouth he explains how important flossing your teeth is. \\
6) A dentist holds the replica of a human mouth he presents how important flossing your teeth is. \\
7) A dentist holds the replica of a human mouth he demonstrates how important flossing your teeth is. \\
8) A dentist holds the replica of a human mouth he communicates how important flossing your teeth is.\\
9) A dentist holds the replica of a human mouth he displays how important flossing your teeth is. \\
10) A dentist holds the replica of a human mouth he highlights how important flossing your teeth is.\\
Input: This is a video of somebody touching wood. \\
Outputs:\\
1) This is a video of somebody tapping wood. \\
2) This is a video of somebody stroking wood. \\
3) This is a video of somebody pressing wood. \\
4) This is a video of somebody handling wood. \\
5) This is a video of somebody patting wood. \\
6) This is a video of somebody brushing wood. \\
7) This is a video of somebody grazing wood. \\
8) This is a video of somebody poking wood. \\
9) This is a video of somebody caressing wood. \\
10) This is a video of somebody gripping wood.\\
Input: This is a video of a group of adults outside dancing. \\
Outputs: \\
 1) This is a video of a group of adults outside whirling. \\
 2) This is a video of a group of adults outside twirling.\\
 3) This is a video of a group of adults outside swaying. \\
 4) This is a video of a group of adults outside partying. \\
 5) This is a video of a group of adults outside getting down. \\
 6) This is a video of a group of adults outside spinning. \\
 7) This is a video of a group of adults outside bouncing. \\
 8) This is a video of a group of adults outside bopping.\\
 9) This is a video of a group of adults outside waltzing. \\
 10) This is a video of a group of adults outside prancing.\\
Input: \textbf{\{input caption\}}\\
Outputs: \\}
\noindent\rule{8.5cm}{0.4pt}

\subsection{T5 generations}\label{subsec:app:t5}

As well as using PaLM to generate hard verb negative captions, we experiment with using  a bidirectional language model, T5-Base~\cite{2020t5}: a 220 million parameter encoder-decoder Transformer. 
It is pretrained on the Colossal Clean Crawled Corpus (C4)~\cite{palm} on a multi-task mixture of unsupervised and supervised tasks, with all tasks being converted into a text-to-text format. 
T5 is trained with a Masked Language Modelling (MLM) loss, similarly to BERT~\cite{devlin-etal-2019-bert}, with minor differences. 
MLM involves masking certain tokens in an input sequence before passing them to the model, and tasking the model with predicting the masked spans. 

As T5 has been trained with a span-mask denoising objective, we use it at inference time in \textit{cloze} form (fill in the blanks) to replace words in captions by targeted masking. 
Specifically, our method consists of the following steps: \\ 
(1) \textbf{Verb Identification:} we start by identifying verbs in text captions, leveraging PoS tagging with NLTK~\cite{bird2009natural}. \\
(2) \textbf{T5 prediction:}
We then replace the verb tokens with a \texttt{[MASK]} token, and feed the masked sentence to T5. We keep the Top-$K$ phrases predicted by the model (with $K = 50$). 
Unlike~\cite{park-etal-2022-exposing}, we do not fine-tune T5 for verb modelling specifically, but rather use it in a zero-shot setting, which we find is sufficient to generate plausible negatives. \\
(3) \textbf{Negatives Filtering:}
The $K$ candidate sentences are then filtered to remove sentences which contain the same verbs as the original caption.

\subsection{Kinetics-verb}\label{subsec:app:kinetics-verb}
In order to assess our method's true verb understanding in the downstream task of action classification, we introduce `Kinetics-verb': a subset of 97 classes from Kinetics-400~\cite{Carreira_2017_CVPR} where we isolate classes that share a \textit{common noun} with another class, but have a \textit{different verb} (and therefore action).  We include the set of 97 classes below:

\noindent [\textbf{hair}: braiding hair, brushing hair, curling hair, dying hair, fixing hair, washing hair, getting a hair cut; \textbf{nails}: doing nails, cutting nails;  \textbf{legs}: waxing legs, massaging legs, shaving legs, stretching leg, swinging legs;  \textbf{hands}: washing hands, shaking hands,  \textbf{arm}: stretching arm, exercising arm, arm wrestling; \textbf{watermelon}: cutting watermelon, eating watermelon;  \textbf{floor}: mopping floor, cleaning floor, sanding floor, sweeping floor;  \textbf{baby}: baby waking up, carrying baby, crawling baby; \textbf{back}: waxing back, bending back, massaging back; \textbf{feet}: massaging feet, washing feet; \textbf{dog}: walking the dog, grooming dog, training dog; \textbf{cake}: eating cake, making a cake;  \textbf{guitar}: strumming guitar, playing guitar, tapping guitar; \textbf{cards}: shuffling cards, playing cards; \textbf{present}: wrapping present, opening present; \textbf{egg}: cooking egg, egg hunting, scrambling eggs; \textbf{shoes}: shining shoes, cleaning shoes; \textbf{pool}: cleaning pool, jumping into pool; \textbf{snow}: biking through snow, shoveling snow; \textbf{rope}: skipping rope, climbing a rope; \textbf{fish}: catching fish, feeding fish; \textbf{eyebrows}: filling eyebrows, waxing eyebrows; \textbf{computer}: using computer, assembling computer; \textbf{tree}:  climbing tree, planting trees, trimming trees; \textbf{car}: driving car, pushing car;  \textbf{golf}: golf chipping, golf driving, golf putting; \textbf{beer}:  drinking beer, tasting beer; \textbf{horse}:  grooming horse, riding or walking with horse; \textbf{paper}: folding paper, ripping paper, shredding paper; \textbf{fire}: extinguishing fire, juggling fire; \textbf{head}: shaking head, shaving head;  \textbf{water}: surfing water, water skiing, water sliding; \textbf{ice}: ice climbing, ice fishing, ice skating; \textbf{basketball}: dunking basketball, dribbling basketball, playing basketball, shooting basketball; \textbf{finger}: drumming fingers, finger snapping;  \textbf{baseball}:  catching or throwing baseball, hitting baseball; \textbf{soccer ball}: juggling soccer ball, kicking soccer ball]

\end{document}